\documentclass[journal]{IEEEtran}
\ifCLASSINFOpdf
\else
\fi
\usepackage{adjustbox}
\usepackage{amssymb}
\usepackage{amsmath}
\usepackage{array}
\usepackage{bbm}
\usepackage{graphicx}
\usepackage{multirow}
\usepackage{rotating}
\usepackage{subfig}
\usepackage{makecell}

\usepackage[ruled]{algorithm2e}
\usepackage{algorithmicx}  
\usepackage{algpseudocode}  
\usepackage{amsmath}

\usepackage[normalem]{ulem}
\useunder{\uline}{\ul}{}

\begin{document}
%
\title{Informative Data Selection with Uncertainty for Multi-modal Object Detection}
%
%
%

\author{Xinyu Zhang,
        Zhiwei Li,
        Zhenhong Zou,
        Xin Gao,
        Yijin Xiong,
        Dafeng Jin,
        Jun Li,
        and Huaping Liu 
\thanks{This work was supported by the National High Technology Research and Development Program of China under Grant No. 2018YFE0204300, and the National Natural Science Foundation of China under Grant No. 62273198, U1964203.}
\thanks{Zhiwei Li and Zhenhong Zou are the corresponding authors.}
\thanks{Xinyu Zhang is with the School of Transportation Science and Engineering, Beihang University, Beijing 100191, China, and with the
State Key Laboratory of Automotive Safety and Energy, Tsinghua University, Beijing 100084, China (e-mail:
xyzhang@tsinghua.edu.cn)}
\thanks{Zhenhong Zou, Dafeng Jin and Jun Li are with the State Key Laboratory of Automotive Safety and Energy, and the School of Vehicle and Mobility, Tsinghua University, Beijing, 100084 China. (e-mails: zouzhenhong@tsinghua.edu.cn, jindf@mail.tsinghua.edu.cn, lijun19580326@126.com)}
\thanks{Xin Gao and Yijin Xiong are with China University of Mining \& Technology, Beijing, China. (e-mails: bqt2000405024@cumtb.edu.cn, bqt2000405025@cumtb.edu.cn)}
\thanks{Zhiwei Li is with the Beijing University of Chemical Technology, Beijing, China. (e-mail: 2022500066@buct.edu.cn)}
\thanks{Huaping Liu is with the Department of Computer Science and Technology, Tsinghua University, Beijing, China. (e-mail: hpliu@tsinghua.edu.cn)}

}

%
%

\markboth{IEEE Transactions on Neural Networks and Learning Systems}
{Shell \MakeLowercase{\textit{et al..}}: Bare Demo of IEEEtran.cls for IEEE Journals}
%



\maketitle

\begin{abstract}
Noise has always been nonnegligible trouble in object detection by creating confusion in model reasoning, thereby reducing the informativeness of the data. It can lead to inaccurate recognition due to the shift in the observed pattern, that requires a robust generalization of the models. To implement a general vision model, we need to develop deep learning models that can adaptively select valid information from multi-modal data. This is mainly based on two reasons. Multi-modal learning can break through the inherent defects of single-modal data, and adaptive information selection can reduce chaos in multi-modal data. To tackle this problem, we propose a universal uncertainty-aware multi-modal fusion model. It adopts a multi-pipeline loosely coupled architecture to combine the features and results from point clouds and images. To quantify the correlation in multi-modal information, we model the uncertainty, as the inverse of data information, in different modalities and embed it in the bounding box generation. In this way, our model reduces the randomness in fusion and generates reliable output. Moreover, we conducted a completed investigation on the KITTI 2D object detection dataset and its derived dirty data. Our fusion model is proven to resist severe noise interference like Gaussian, motion blur, and frost, with only slight degradation. The experiment results demonstrate the benefits of our adaptive fusion. Our analysis on the robustness of multi-modal fusion will provide further insights for future research.
\end{abstract}


%
\IEEEpeerreviewmaketitle

\section{Introduction}
%
%
%
%
\IEEEPARstart{R}{ecent} success in deep learning has contributed much to computer vision, such as semantic segmentation, object detection, and object tracking. However, for some application areas like autonomous driving, there are more requirements for vision models\cite{Feng2021DeepMO}. Though current models can perform well on most tasks, they have a limitation on dirty data and fail to meet the practical standard of industrial application\cite{Bijelic2018RobustnessAU,Bijelic2020SeeingTF,Hnewa2021ObjectDU}. For instance, when a self-driving vehicle (SDV) runs on the road, complex traffic scenarios, unpredictable weather conditions, and potential sensor failure can lead to pattern shifts and inaccurate object recognition. Therefore, robustness and generalization are gradually brought into focus in model development. In the following, we explain why multi-modal learning and adaptive selection of informative data are important for improving model generalization.

\begin{figure}[t!]
    \centering
    \includegraphics[width=1\linewidth]{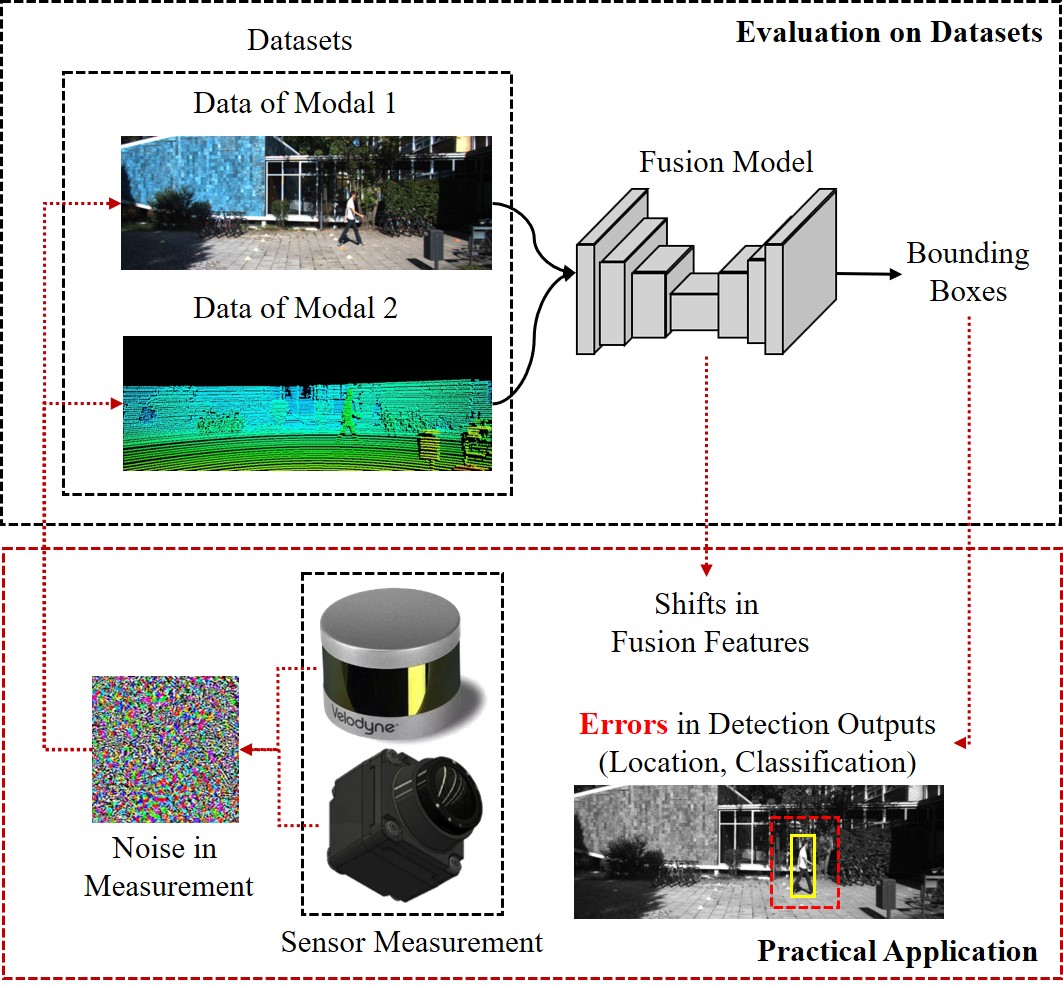}
    \caption{Illustration of the measurement noise in perception models. Tests on datasets take fixed data points as input and generate statistical conclusions. While in practical application, various environment and hardware conditions produce potential noise in the measurement. It will result in a shift in multi-modal features and the errors in the final output.}
    \label{illustration}
\end{figure}

Many methods have been proposed to solve the problem, varying from the hardware side like using more elaborate sensors and the software side like adaptive algorithms. Some algorithms estimate the correct features for different data, including data augmentation\cite{Ma2021PointDropIO}, domain adaptation\cite{Zhuang2021ACS,Yi2020AdaptiveWS}, feature enhancement\cite{Bijelic2020SeeingTF}, noise estimation\cite{Yang2020LaNoisingAD}, and so on. They avoid the affects from noise in data by grabbing the invariant features or modifying the training data distribution. However, most of these methods are developed on single-modal models, which rely on the specific sensor measurement essentially. As illustrated in the Fig.\ref{illustration}, generally we train and evaluate our models on the given datasets. Due to the limited amount of data, the statistical results on the datasets cannot reflect the true performance in reality, especially when sensors meet error beyond the cases in datasets, namely the out-of-distribution(OOD) problem. Therefore, redundancy in the data is also a vital aspect in practical deployment.

For the reasons above, multi-modal fusion methods have gained attention in recent years\cite{Feng2021DeepMO,Baltruaitis2019MultimodalML}. SDVs are equipped with different sensors for more complete and precise perception. On one hand, multi-modal sets can provide complementary measurements, like cameras record the colors and textures, LiDARs provide the 3D structure of objects, and Radars observe the velocity of moving targets. On the other hand, multi-modal fusion can provide redundant information for stable recognition. Different sensors have specific working conditions, which means they all will potentially fail in some environments. For example, in the dark or foggy weather or with sharp illumination changes, the images from the camera may fail to recognize objects. As for LiDARs, particulate matters in the air can influence the LiDAR imaging in rainy, snowy, and sandy weather. With data fusion, sensing systems can avoid severe crash. Therefore, we mainly discuss the adaptive strategy of multi-modal fusion for SDV perception in this paper, especially on the object detection task.

There has been much research in the multi-modal object detection of autonomous driving since the KITTI benchmark was released in 2012\cite{Geiger2012AreWR}. It is also recognized as a potential approach to realize robust detection. After that, various datasets and approaches have been proposed to accelerate the development of this community and achieve higher recognition accuracy as well\cite{Caesar2020nuScenesAM,Sun2020ScalabilityIP}. In the early stage, model-based methods used bagging methods for result fusion\cite{Oh2017ObjectDA,Xu2018PointFusionDS}. They generally have individual pipelines to process different data and merge the bounding boxes to generate the final results. Latest data-driven methods mainly applied feature fusion (boosting or stacking) for a more profound information mixture that fuses multi-modal data in the feature extraction or Region of Interest (ROI) generation stage\cite{Bijelic2020SeeingTF,Liang2018DeepCF,Qi2018FrustumPF}. 

However, existing fusion methods focus on the quantified scores in standard vision benchmarks, while few contribute to the robustness or generalization of fusion\cite{Feng2020LeveragingUF}. Models that can adapt to different inference datasets are critical for real-world applications. They can aggregate useful information from the high-dimensional feature space, thus avoiding the effect of noise on the results. It was pointed out a multi-modal model seek a balance between fitting data and generalization\cite{Zou2021ANM}. Its performance will meet a bottleneck without proper information reduction in fusion channels. Due to the same reason, it is more challenging to extend the training data via multi-modal data augmentation. In addition, when a multi-modal model has not been trained well as in Fig.\ref{illustration}, it will generate greater variance with dirty multi-modal data. As a result, multi-modal fusion does not essentially guarantee an incrementation in performance and robustness. In other words, multi-modal models are not always effective in identifying and exploiting information in diverse data.

\begin{figure}[t!]
    \centering
    \includegraphics[width=1\linewidth]{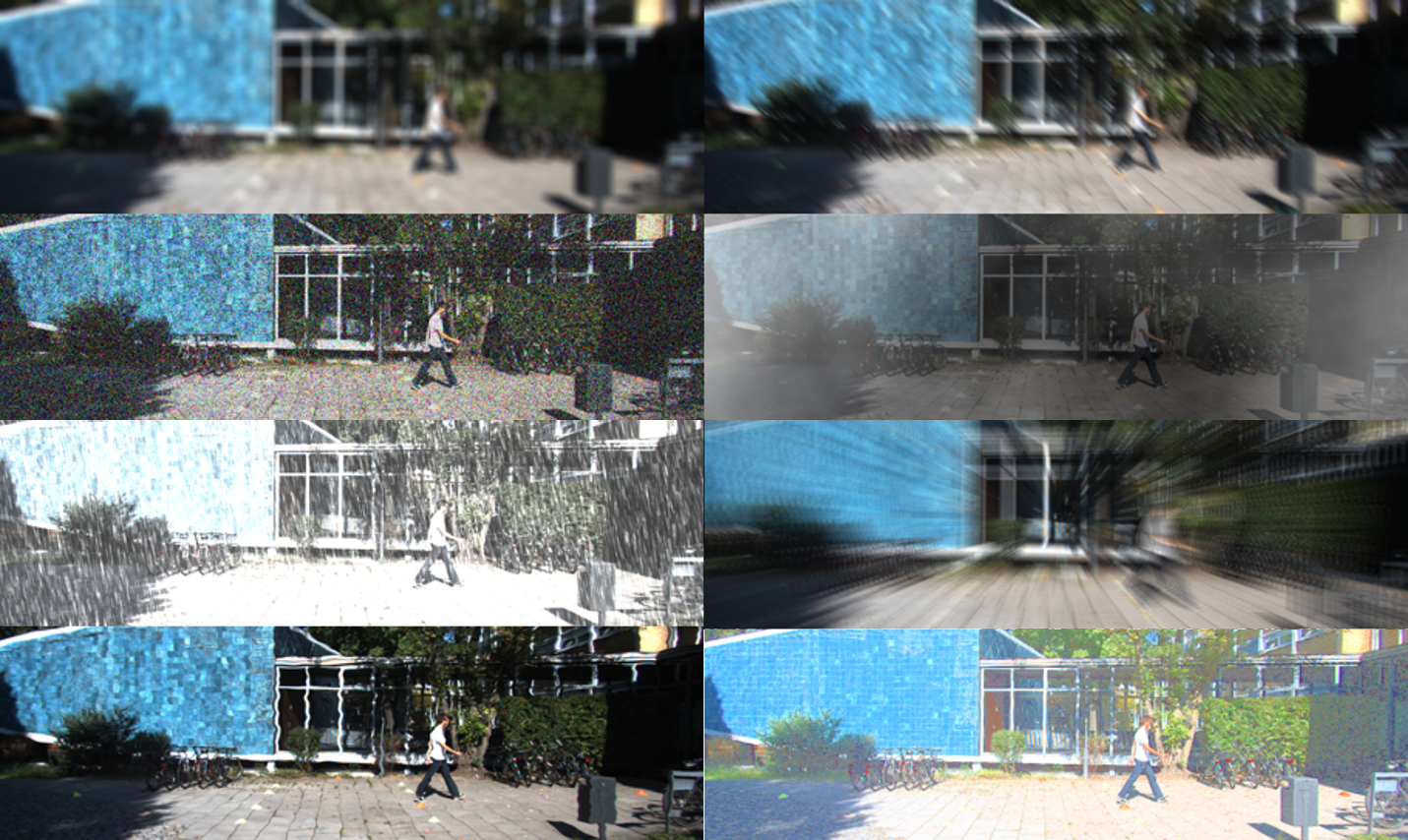}
    \caption{Visualization of the simulated noise\cite{Michaelis2019BenchmarkingRI}. From top left to bottom right are eight common noises in nature.}
    \label{noisesim}
\end{figure}

Different from those learning-based feature fusion models, we claim that adaptive fusion models should select informatively features from multi-source data and avoid noise in them. It can be viewed as one form of data dimensionality reduction, which is expected to be explainable. The amount of information is measured by entropy, so it can be equivalent to the calculation of the feature distribution. Since it is difficult to directly calculate the amount of information or entropy, we can choose to reflect the uncertainty of feature distribution to learn to characterize the amount of information. Driven by this idea, we propose an adaptive fusion method with a result fusion architecture. Considering that different modal data require specific operators and parameter optimization in feature processing, we adopt a loosely coupled network architecture, which is general but practical. Multi-modal data are fed to individual pipelines that are connected in the boxes filtering stage. Then we apply decision-level fusion by fusing the box proposals in the improved Non-Maximum Suppression (NMS). We will describe why our simple fusion strategy can filter information in data.

To select informative results and achieve reliable fusion, we introduce uncertainty quantification into our model. Proper uncertainty quantification indicates the prediction deviation of the model, therefore, it has been viewed as a potential approach towards the interpretable neural network and an emerging method in autonomous driving\cite{Kendall2017WhatUD,Feng2021DeepMO}. In our model, we predict the uncertainty, as the inverse of information amount, for each data point. Through joint training on multi-modal data, our model learns a universal uncertainty measurement that can be used as the boxes filter index in NMS. To demonstrate the benefits of our design, and explore the noise affect in fusion models as well, we have evaluated the models on the KITTI 2D object detection dataset. Point clouds and RGB images were progressively perturbed to simulate multi-level dirty data. Then, we conducted experiments with both raw data and dirty data. For clean data, our fusion model achieved sub-optimal but competitive results, with only 0.36 mAP lower than Depth-model, while 4.24 mAP higher than RGB-model. As for dirty data, our achieved 51.61 mAP higher than RGB-model and 34.20 mAP higher than Depth-model on average. Our main contributions can be concluded as:
\begin{itemize}
    \item We explored the influence of multi-level noise on LiDAR point clouds and Camera RGB images and reveal the attenuation law for object detection task;
    \item We proposed a universal fusion model with informative data selection, which can be implemented with different modal data and fuse their predictions adaptively;
    \item We conducted sufficient experiments on the KITTI dataset, that demonstrate our model has strong robustness and generalizes to noisy data beyond the train set.
\end{itemize}

In the following sections, we first review the recent progress in anti-noise object detection, multi-modal fusion, information in fusion, and uncertainty for computer vision. Then, we introduce our proposed model from baseline, uncertainty modeling, fusion step, and implementation. After that, we detail the experiment process, including data pre-processing, noise simulation, results and analysis.

\section{Related Work}

\subsection{Anti-Noise Object Detection}
Noisy data can mislead detection models because their object features are out of the distribution of the model fitting domain. Therefore, it is practical to extend the training set to provide more diverse data, or create feature filters to regularize noisy features. Data augmentation is one of the most common approaches for the former. PointDrop\cite{Ma2021PointDropIO} learns to drop some key points as features, thus generating more challenging point samples for training. Ofori-Oduro et al. used antibodies generated using Artificial Immune Systems in training\cite{OforiOduro2020DataAU}. Loh et al. proposed to collect data under different illumination conditions to enhance the model's robustness\cite{Loh2019GettingTK}. Michaelis et al.\cite{Michaelis2019BenchmarkingRI} provided a benchmark to simulate multiple noises in a natural environment that is presented in the Fig.\ref{noisesim}. Besides, domain adaptation is another practical method. Transferring learned parameters(knowledge) among datasets can aggregate features distributions in different data domain\cite{Zhuang2021ACS}. Khodabandeh et al. proposed to use noisy-label in training to enhance the generalization\cite{Khodabandeh2019ARL}. Instead, researcher also concerns that whether we can directly extract and enhance related features. Bijelic et al. proposed a multi-sensor model and mix features in multiple levels for mutual activation\cite{Bijelic2020SeeingTF}. Others also try to estimate the noise from the opposite perspective. Yang et al. built a model to estimate the accuracy of Laser measurement under foggy conditions\cite{Yang2020LaNoisingAD}, and Tian et al. quantified the uncertainty level of features for adaptive fusion\cite{Tian2020UNOUN}. But most of them are developed on single-modal models, which indicates that they would fail when the sensor meets severe fault.

\subsection{Multi-modal fusion for object detection}
\subsubsection{Multi-modal object detection}
To date, several studies have investigated multi-modal fusion for 2D and 3D object detection. Frustum PointNets\cite{Qi2018FrustumPF} extract the 3D bounding frustum of an object by extruding 2D bounding boxes from image detectors. PointFusion\cite{Xu2018PointFusionDS} combines a CNN and a PointNet\cite{Qi2017PointNetDL} architecture respectively to process images and raw point clouds then predict 3D boxes. PointPainting\cite{Vora2020PointPaintingSF} projects LiDAR points into the output of an image-only semantic segmentation network, and appends the class scores to each point. All these fusion methods of RGB and LiDAR achieve high average precision on the benchmarks, however, the coupling or interrelation of two modalities will cause the whole system to fail easily once part of the sensors break down. Besides, the methods above only provide a deterministic predict result, making it risky to carry out in the real application.
\subsubsection{Adaptive fusion}
Several new studies have proposed self-adaptive techniques in computer vision. Therefore, the robustness of those tasks can be improved to some extent. Adaptnet\cite{Valada2017AdapNetAS} uses a convoluted mixture of deep experts(CMoDE) fusion techniques to learn features from complementary modalities and spectra. And SSMA\cite{Valada2019SelfSupervisedMA} further proposes a self-supervised model adaptation fusion mechanism and a segmentation architecture termed AdapNet++ to improve the robustness of the system. UNO\cite{Tian2020UNOUN} proposes a Noisy-or Gate, then the model tends to accept the more reliable modal data. The self-adaptive network can recognize potential degradation of the input data such as rain, fog, or image blur, however, they only validate their scheme in a simulation environment, making it less conceivable to real world application.

And all of these methods above are designed for semantic segmentation task, which is less complicated than object detection task. Choosing Smartly\cite{Mees2016ChoosingSA} provides an adaptive fusion technique for multi-modal object detection in changing environments, which predicts weights for different modalities online based on CNN experts. Zhang et al.\cite{Zhang2021ProgressiveMC} propose the PMC method to adaptive generate and fuse different modal data even when they are lost, to achieve cross-domain multi-modal learning. Zhao et al.\cite{Zhao2021AdaptiveCM} propose to apply a gate module for adaptive feature fusion in learning. Kim. et al.\cite{Kim2018RobustDM} propose a gated information fusion network for robust deep multi-modal object detection using separate CNN and SSD(Single Shot Detector)\cite{Liu2016SSDSS} as backbone. The network learns weights from input feature map from each modality in the presence of the modalities degraded in quality. Their following work proposes to design a specific loss function and feature fusion module to avoid noisy data in training\cite{Kim2019OnSS}. Different from those feature fusion models, Lee et al. proposed a late(result-level) fusion method named DBF\cite{Lee2021DBFDB}, that applied fuzzy state estimation for multiple detectors outputs and fused them accordingly for a single image. Inspired by their work, we consider late fusion for adaptive multi-modal fusion. 

As presented in the papers\cite{Bijelic2018RobustnessAU,Zou2021ANM} that conducted several noisy data experiments on the KITTI dataset, existing methods have some problems when apply to real scenarios given that the network is hard to train so that weight may be apt to rely more on those easy-to-learn modals than on harder one. Furthermore, in terms of the back propagation, due to the lack of noise data in the training data, it is difficult to completely cover the existing noise types. Its distribution is always unbalanced relative to that in the real world. As a result, the parameter optimization after back propagation is not suitable for the case that the data contains interference, namely the OOD problem. Therefore, the fusion must be done in a way that is independent of the data distribution. The information-driven data selection proposed in this paper is implemented through a non-parameterized NMS process to avoid the problem of data distribution.

\subsection{Information in multi-modal fusion}
Information theories are proposed by C.Shannon based on the information quantification models\cite{shannon2001mathematical}. It measures the amount of value transferred by signals. But counting information in deep networks is difficult because of the high-dimension data and models. MacKay et al. have discussed modeling a neuron or a network as a channel\cite{mackay2003information}. N. Tishby et al. proposed the information bottleneck theory (IB) with a variational principle to reflect on signal processing problems\cite{tishby2000information,tishby2015deep}. But these researches mainly focus on simple networks. Belghazi et al.\cite{belghazi2018mutual} proposed a mutual information measurement method, that can be applied in simple informative data selection models\cite{zou2020mimf}. A similar conclusion can be found in this paper\cite{Deng2019DomainAV}, which solved the domain adaptation problem via feature selection. Zou et al.\cite{Zou2021ANM} tried to reveal the principle in deep multi-modal networks with information communication models, while they lack compelling modeling, inference, and validation. All in all, we still lack a simple, efficient, and interpretable information fusion method. There are other works like \cite{xu2020u2fusion} leverage self-defined information measurement in vision tasks, but they mainly focus on low-level tasks like image recovery or super-resolution, which may not fit high-level tasks.

\subsection{Uncertainty estimation in computer vision}
Recently, more attention focus on the provision of interpretability, beyond this, uncertainty estimation is one of the most important partition. General uncertainty estimation on computer vision including classification and regression. Gal et al.\cite{Gal2016UncertaintyID} propose a framework to combine aleatoric uncertainty and epistemic uncertainty. They apply the framework to segmentation tasks and achieve new state-of-the-art results on depth regression and semantic segmentation benchmarks. UNO\cite{Tian2020UNOUN} presents an uncertainty-aware fusion scheme and an additional data-dependent spatial temperature scaling method to complement the uncertainty estimation in semantic segmentation. There are also several techniques  for uncertainty estimation in object detection. He et al.\cite{He2019BoundingBR} provide a bounding box regression with KL loss and substitute softer-NMS for traditional NMS so that they achieved more accurate object localization. Choi et al.\cite{Choi2019GaussianYA} modified parameters of YOLO network to gaussian distribution and use loss attenuation to compute the uncertainty of bounding box, and they also improve the mean average precision on the benchmark dataset. Similarly, Lee et al.\cite{Lee2020LocalizationUE} purpose a method named Gaussian-FCOS which uses an anchor-free backbone and even achieves better performance. Kowol et al.\cite{Kowol2021YOdarUS} also present an uncertainty-based strategy with camera and Radar with YOLOv3\cite{Redmon2018YOLOv3AI} as baseline and use gradient boosting to make decision. They demonstrated that the strategy is effective in night scenes compared with single sensor baseline. Feng et al. progressively provide three works on 3D object detection based on uncertainty estimation, from pure point clouds to multi-modal fusion\cite{Feng2018TowardsSA,Feng2019LeveragingHA,Feng2020LeveragingUF}. With precise estimation, they can avoid the uncertain information from noisy samples in training, and aid the prediction for test data. Russell et al.\cite{Russell2019MultivariateUI} show that for the visual tracking problem, accurate multivariate uncertainty quantification can have a great impact on performance for both in-domain and out-of-domain evaluation data.

Most of the models above avoid specific types of noisy data. But most of these techniques do not consider either multi-modal situations or robustness of the model under various noise attacks. For example, Gaussian YOLOv3\cite{Choi2019GaussianYA} is developed for single-modal data, UNO\cite{Tian2020UNOUN} is evaluated on semantic segmentation tasks, Feng\cite{Feng2020LeveragingUF} and Zhang\cite{Zhang2021ProgressiveMC} ignore many other types of noise data from environment. Hence, we intend to synthesize the advantages of those techniques in an uncertainty-aware multi-modal object detection model, and evaluate it with common types of dirty data.

\begin{figure}[t!]
    \centering
    \includegraphics[width=1\linewidth]{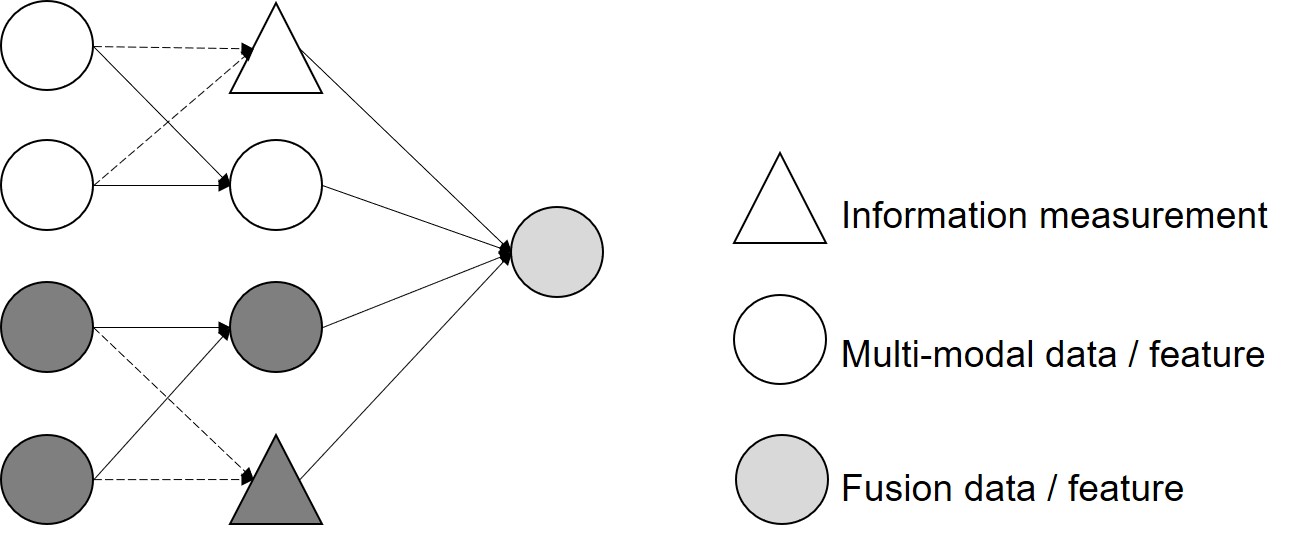}
    \caption{A simple model of informative data selection in fusion. When multi-source data or features are combined, the model should measure their information amount as a reference. In the case of noisy-data detection, it is critical to pick out those related element and filter out the noise.}
    \label{selection}
\end{figure}

\section{Method}
In the following, we first formulate the noisy-data object detection problem, and describe our baseline fusion model. To improve the robustness and generate informative data selection fusion strategy, we build an uncertainty-aware model by estimating the aleatoric uncertainty of each bounding box of each modal via loss attenuation. Then it will be applied for boxes filtering in the NMS stage. To balance speed and accuracy, we choose a light-weight one-stage object detection model as our baseline, while our method should be easily extended to other classical object detection models.

\subsection{Problem Statement}
Noisy-data object detection aims to locate and classify targets from data that is disturbed with natural noise. Generally an object detection model $\mathcal{D}(\cdot)$ take an image, point clouds, or a group of multi-modal data $X=\{X_1,X_2,...\}$ as input, where subscripts indicate the modality. Then it returns the expected coordinates $\{x,y,w,h\}$ and categories $\{c\}$ of the targets. For noisy-data detection, we assume that clean data will be measured by noisy function $\mathcal{F}(\cdot)$. Our goal is to minimize the recognition deviation while maximizing the influence of noise as much as possible:
\begin{equation}
    \min_{\mathcal{D}}\max_{\mathcal{F}}\mathcal{L}(\mathcal{D}(\mathcal{F}(X)),\{x,y,w,h,c\})
    \label{definition}
\end{equation}

As we approach the minimum, we can guarantee the generalization and robustness of our detection models over the most severe noise. Specifically, for multi-modal fusion models, to simplify the problem, we assume that $X$ contains at least two modal data, namely two types of sensors at least. In this paper, we focus on LiDAR point clouds and camera RGB images. In addition, noise will only be added to one modal data in a single experiment, to avoid the worst-case scenario where all sensors fail and nothing will work. In the case of a multimodal model, the Expr.\ref{definition} becomes:
\begin{equation}
    \min_{\mathcal{D}}\max_{\mathcal{F}}\mathcal{L}(\mathcal{D}(\mathcal{F}(X_{l}),\mathcal{F}(X_{c})),\{x,y,w,h,c\})
    \label{definition1}
\end{equation}
where $X_l$ is LiDAR data and $X_c$ is camera data. For our informative data selection sub-problem, it will be revised as,
\begin{equation}
    \min_{\mathcal{D}}\min_{\mathcal{S}}\max_{\mathcal{F}}\mathcal{L}(
        \mathcal{S}(
            \mathcal{D}(\mathcal{F}(X_{l})),
            \mathcal{D}(\mathcal{F}(X_{c}))
            ),
        \{x,y,w,h,c\})
    \label{definition2}
\end{equation}
where $\mathcal{S}(\cdot)$ is the selective fusion function as shown in Fig.\ref{selection}.

\begin{figure*}[t!]
    \centering
    \includegraphics[width=1\linewidth]{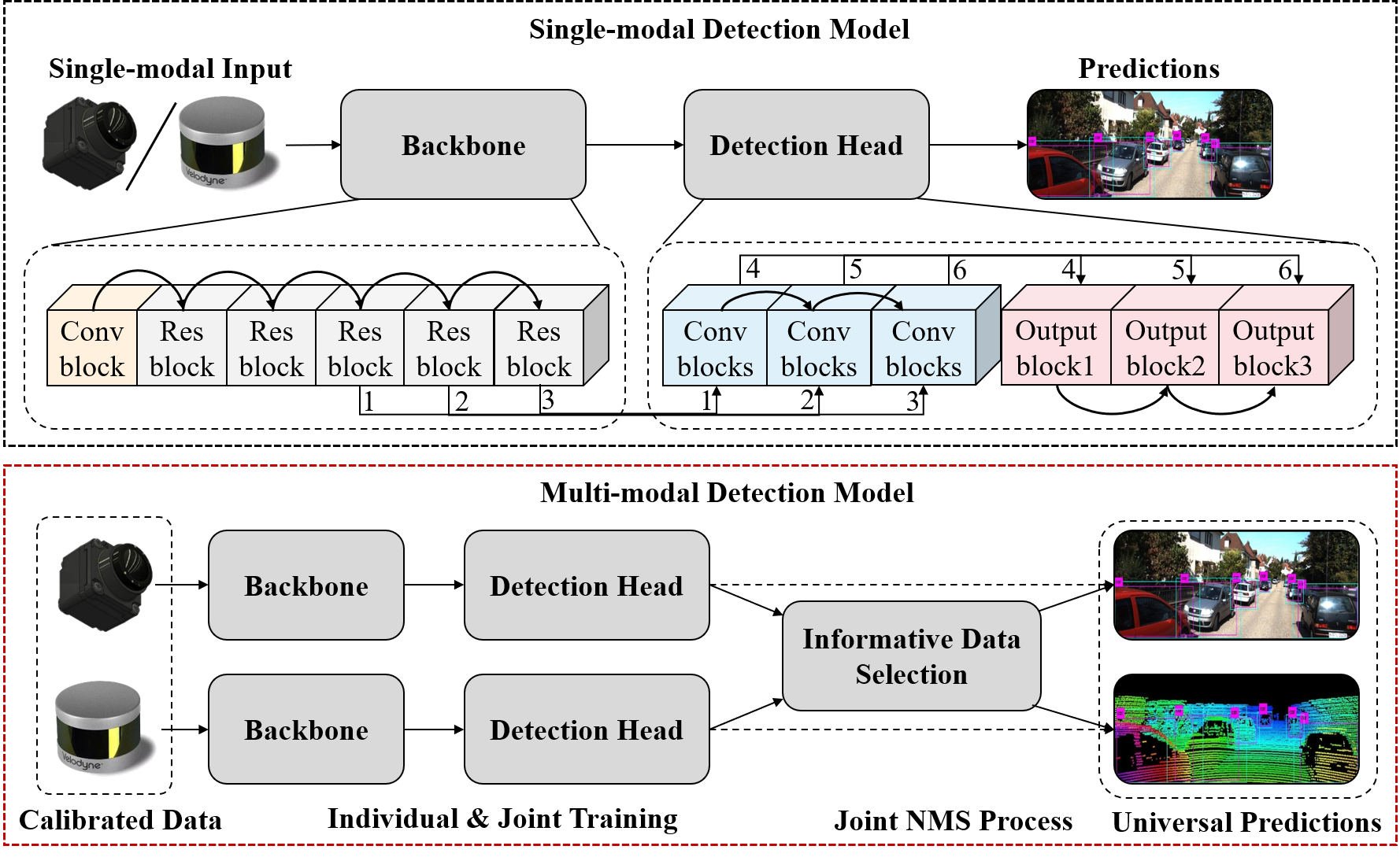}
    \caption{Architecture of the proposed 2D object detection model. The top sub-figure presents the overview of YOLOv3 network\cite{Redmon2018YOLOv3AI}. Details of the network layers refer to the original paper. The bottom sub-figure shows our loosely-coupled multi-modal detection model with a join NMS for universal prediction in the 2D space. Besides, our model can be easily implemented with other backbones like SSD\cite{Liu2016SSDSS}.}
    \label{arch}
\end{figure*}

\subsection{Detection Baseline}
The YOLOv3\cite{Redmon2018YOLOv3AI} is a modified version of the YOLO series models, achieving practicality in terms of accuracy and speed. The structure of YOLOv3 has been shown in the Fig.\ref{arch} above, consisting mainly of the DarkNet-53 backbone network and the three-branch decoder. The backbone network is composed of residual blocks and convolution blocks, while each decoder branch mainly contains the convolution blocks. A convolution block comprises a 2D convolution layer, a batch normalization layer, and a Leaky ReLU activation. A residual block comprises various residual units, and each of them has two convolution blocks and a skip-layer concatenation. Decoders process the feature maps transmitted by the former blocks and the previous branch, which will be up-sampled to realize multi-scale detection. These designs improve the detection robustness and speed of YOLOv3. To apply YOLOv3 for LiDAR point clouds, we project points onto a 2D plane as the camera image, and process it as a 2D depth image.

\subsection{Multi-modal Fusion}
As mentioned above, in most existing multi-modal fusion networks, the modalities are heavily coupled and do not consider any safety-critical environment. To weaken the interfere for fear that the modality will harm each other, we use parallel pipelines and use late fusion to combine the final result. The overview architecture are presented in the Fig.\ref{arch}. RGB images and the projected depth images will be calibrated in space and size. Then they will be fed to individual pipelines. Though we apply YOLOv3 for both modalities, they also accepted different combinations of models, whether 3D or 2D, one-stage or two-stage. Besides, a potential scenario is to establish communication between the two models by fusing two modal features in the middle of the models\cite{Feng2020LeveragingUF,Zou2021ANM}. Says it may get better characteristics, but in our exploring uncertainty-aware detection model, we don't do this in order to avoid the model non-convergence issue and two modal data uncertainty interfere with each other. For the basic fusion model, we can simply fuse all the proposed bounding boxes and filter them in the joint NMS process, or apply NMS for each modality and conduct a second selection for them afterward. In the final version, this process will be revised to the uncertainty-aware multi-source NMS. The fusion model universally outputs all boxes onto the image 2D plane. Because of our loosely coupled design, the proposed fusion model is compatible with more different modalities and tasks, even with the addition of uncertainty fusion factors. But we will take LiDAR point clouds and camera images as example in our experiments.

\subsection{Informative Data Selection with Uncertainty}
Considering that it is difficult to calculate an accurate amount of information in data, we leverage a related statistic index uncertainty as the alternative. Both information and uncertainty describe the credibility of data to the target object category and locations, while uncertainty represents the inverse meaning of information or mutual information. In our fusion strategy, we mainly consider aleatoric uncertainty, since the noises in each modality caused by sensor failure or extreme weather can be better explained by aleatoric uncertainty. Generally speaking, aleatoric uncertainty can be interpreted as noise or vague in data or a label that is hard to fit.

Though aleatoric uncertainty cannot be eliminated by adding more training data, it could be reduced with additional features or views. Therefore, our strategy can also be interpreted as a type of method to reduce aleatoric uncertainty using multiple modalities. To estimate the aleatoric uncertainty of each object in each modality, we apply the loss attenuation that integrates uncertainty estimation function in the training loss, and optimizes them together with neural networks. For object detection, we focus on the coordinates regression and classification. The traditional loss function is:
\begin{equation}
    \mathcal{L}_{NN}(\theta)=\frac{1}{N}\sum_{i=1}^N ||y_i-f(x_i)||^2
    \label{detloss1}
\end{equation}
where $NN$ is the neural network, $N$ is the number of samples, $\{x_i\}$ and $\{y_i\}$ are the input and target output for $(x,y,w,h)$ in the Eq.\ref{definition}. To recognize the uncertain predictions in fusion, we expect the model can assign high uncertainty to inaccurate results and low for the rest. Then the Eq.\ref{detloss1} should be redesigned as\cite{Choi2019GaussianYA}:
\begin{equation}
    \mathcal{L}_{NN}(\theta)=\frac{1}{N}\sum_{i=1}^N \frac{1}{2\sigma(x_i)^2}||y_i-f(x_i)||^2+\frac{1}{2}log\sigma(x_i)^2
    \label{detloss2}
\end{equation}
where $\sigma(\cdot)$ is the variance estimation function. In particular, in the case of a single target, the model may output 0 to N prediction boxes. In the training stage, the optimization of this loss function can constrain the prediction result of the model to be as close to the real value as possible, and the variance between the prediction boxes will not be too large. In the inference stage, based on the prediction variance and error obtained by this publicity, the quality of the prediction frame can be judged, so as to carry out effective fusion. This method and conclusion can be easily generalized to the case of multiple objectives. With the loss attenuation, the model is expected to predict proper uncertainty for boxes. 

\begin{figure}[t!]
    \centering
    \includegraphics[width=1\linewidth]{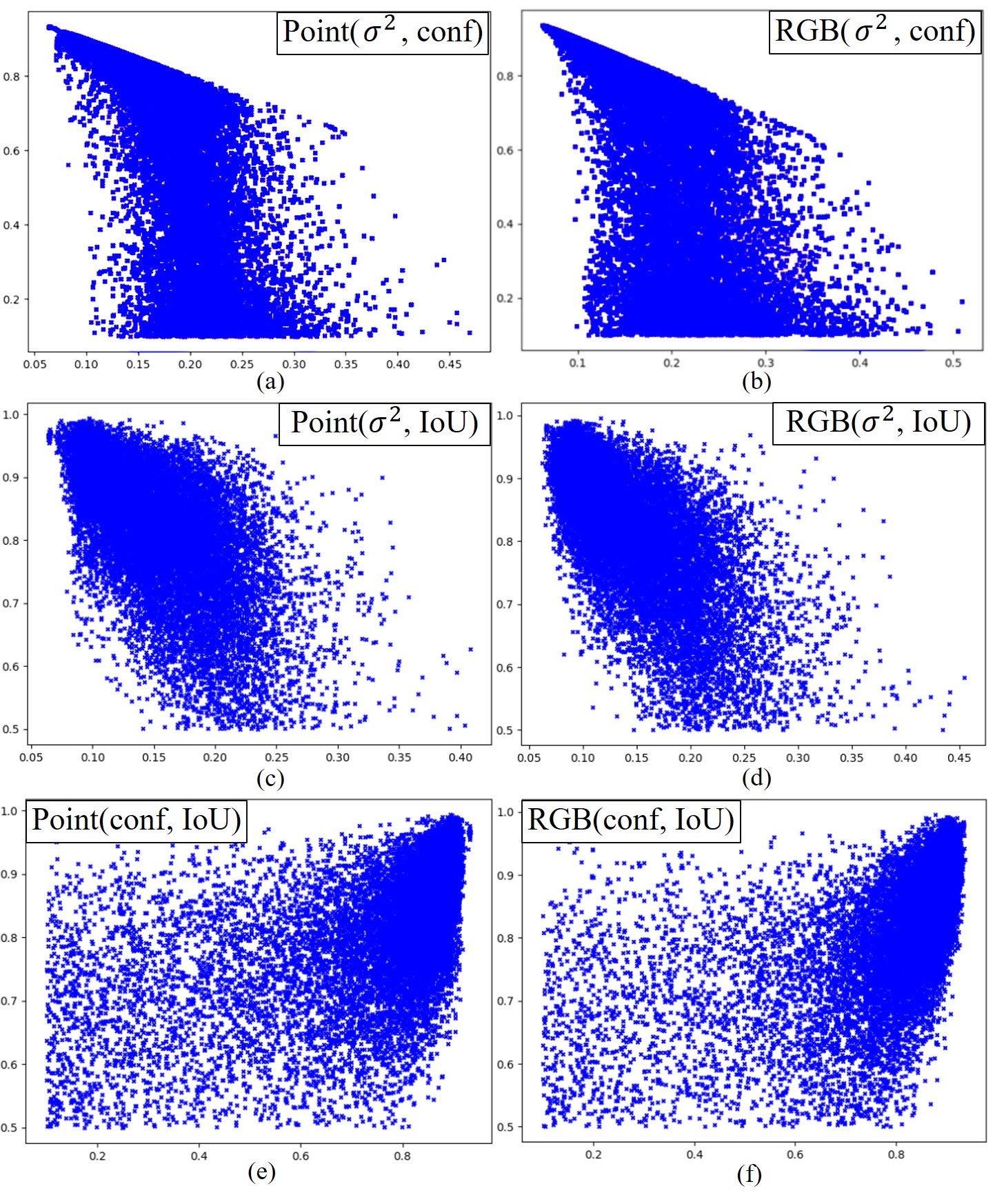}
    \caption{The validity of uncertainty estimation. We visualize the relationship among uncertainty(variance $\sigma^2$), classification confidence, and localization accuracy IoU above. All of these are computed for point-based and image-based models separately as shown in the six sub-figures. On a scatter plot, the degree to which all points converge to a ray from the origin indicates the degree of regression of these points. The degree to which these points are correlated. Therefore, we believe that we can intuitively show the correlation between confidence, uncertainty and IoU three commonly used indicators.}
    \label{valid}
\end{figure}

Following the setting of Gaussian-YOLOv3\cite{Choi2019GaussianYA}, we extend the prediction to $(\mu_x, \Sigma_x, \mu_y, \Sigma_y, \mu_w, \Sigma_w, \mu_h, \Sigma_h)$, where $\mu,\Sigma$ indicate the mean and variance of the target elements. Then the mean should converge to the ground truth, while smaller variances are predicted for accurate boxes and larger variances are predicted for inaccurate boxes. Then the uncertainty can be indicated by variance. Due to the penalty during training, the samples with high uncertainty tend to generate high variance during optimization. Additionally, to fit the prediction mode of YOLOv3, we need to conduct sigmoid function for each value and scale them to the range of (0,1). 

In our experiment, we examine the validation of uncertainty estimation for point clouds model and RGB model on the KITTI 2D object detection dataset. The relationship among confidence(conf), uncertainty($\sigma^2$), and localization(IoU) are visualized in the Fig.\ref{valid}. In the figures, the more linearly correlated the distributions of two variables are, the more correlated, or interpretable, they are. In Fig.\ref{valid}, (a) and (b) present the joint distribution of $\sigma^2$ and conf. Their distributions are not so closer to linear, indicating that their uncertainty are less related to confidence. Compared with this group, it is more interpretable for (IoU, $\sigma^2$) and (IoU, conf). The data point in RGB sub-figures are more concentrated than in points sub-figures. That means our uncertainty estimation fits RGB models better than point cloud models. For the relationship of (IoU, $\sigma^2$) and (IoU, conf), the figures also indicate that the correlation between confidence and positioning accuracy is not significant as uncertainty. To sum up, it is reasonable to fuse and optimize the candidate boxes using uncertainty.

In this way, we model every bounding box in each object with an uncertainty value indicating the quality of the object:
\begin{equation}
    Bbox_i \sim N(\mu_i, \Sigma_i^2)
    \label{detloss3}
\end{equation}

Eq.\ref{detloss3} above is applied for the four elements $\{x,y,w,h\}$. As for our multi-modal model, we have individual pipelines for each input data and correspondingly individual predicted boxes for them, that means we need to model the uncertainty in each pipelines. Because the uncertainty(variance), only reflects the properties of the boxes, we can think of these estimates as belonging to the same scale with similar meanings.

For optimization, Eq.\ref{detloss2} is revised as follows:
\begin{equation}
    \mathcal{L}_x = -\sum_{i=1}^W\sum_{j=1}^H\sum_{k=1}^K\lambda_{ijk}log(P(x_{ijk}^{GT}|\mu_x(x_{ijk}),\Sigma_x(x_{ijk}))+\epsilon)
    \label{detloss4}
\end{equation}
\begin{equation}
    \lambda_{ijk}=\frac{(2-w^{GT}\times h^{GT})\times \delta_{ijk}^{obj}}{2}
\end{equation}
where $K$ is the number of anchors, and $\lambda_{ijk}$ is set as a penalty coefficient of weight. $P(\cdot)$ is an expected distribution probability function for boxes. Generally we take Gaussian distribution here. $\delta$ represents a gate between prediction and anchors. In our experiment, $\delta=1$ when the IoU is over 0.3, otherwise it is 0. The Eq.\ref{detloss4} is also applied for the four elements $\{x,y,w,h\}$ according to the specific loss function. In the setting of our result-level fusion, we predict proposed bounding boxes from two pipelines and fuse them in single NMS module, though we optimize the Eq.\ref{detloss4} individually for two pipelines output. However, uncertainty estimation is not sufficient for adaptive fusion, we further design an uncertainty-aware multi-source NMS algorithm to achieve the goal.

\subsection{Data Selection in NMS}
We finally integrate the selection process in the improved-NMS. NMS can filter bounding box according to IOU and classification score. But the original algorithm only considers classification score and ignore the accuracy of localization. Typically, when considering model uncertainty, the classification score generated by softmax layer are probable to overestimate the score. Besides, lower-scored boxes may have higher localization confidence. To integrate uncertainty and combine the entire results of each modality in our model, softer-NMS\cite{He2019BoundingBR} has been proposed to substitute NMS. It calculates the weighted average of all based on box-level aleatoric uncertainty and update the localization parameters for prediction. For example, $x_1$ is updated by:
\begin{equation}
    x_{1,i} = \frac{\sum_j x_{1,j}/\sigma_{x_{1,j}}^2}{\sum_j 1/\sigma_{x_{1,j}}^2} \ \ \ \ \ \ s.t.IoU(b_i,b)>0.5
\end{equation}
where $\sigma_{x,i}^2$ is aleatoric uncertainty of bounding box, and $b$ indicates the predicted boxes. All eight parameters from two pipelines will be updated with the same approach.

In the case of multi-source fusion, we have predictions from multiple pipelines. If we mix the predictions of multiple modalities directly, we will ignore the pattern correlation across different modalities, and consistency within each modality respectively. Compared with general NMS and other fusion methods, that have different distributions for different datasets, our fusion strategy keeps the high consistency over different modality data and multi-source data, and significant relationship with localization and classification. Therefore, given two threshold $t_1$ and $t_2$, we can classify the relationship between the predictions of the two modalities $A,B$ into three cases:
\begin{itemize}
    \item Case1: when $IoU(A,B) \in [t_2,1]$, the area is activated by two modal data with high confidence;
    \item Case2: when $IoU(A,B) \in [t_1,t_2)$, the area exists confusing patterns from different modalities;
    \item Case3: when $IoU(A,B) \in [0,t_1)$, different modal data detect objects in different areas that are not correlated.
\end{itemize}
These two parameters can obtain an optimal value through experiments. The reference values we provide are 0.3 and 0.5.

\begin{algorithm}[b!]  
    \caption{Uncertainty-aware Multi-source NMS}  
    \label{NMS}
    \KwIn{$\mathcal{B}_r=\{b_1,...,b_n\}$: boxes from RGB images; 
    $\mathcal{S}_r=\{s^2_1,...,s^2_n\}$: confidence of $\mathcal{B}_r$; 
    $\mathcal{C}_r=\{\sigma^2_1,...,\sigma^2_n\}$: variance of $\mathcal{B}_r$;
    $\mathcal{B}_p=\{b_1,...,b_m\}$: boxes from point clouds; 
    $\mathcal{S}_p=\{s^2_1,...,s^2_m\}$: confidence of $\mathcal{B}_p$; 
    $\mathcal{C}_p=\{\sigma^2_1,...,\sigma^2_m\}$: variance of $\mathcal{B}_p$.
    }  
    \KwOut{$\mathcal{D}$: the final set of detections.}
    $\mathcal{D}\gets\{\}$, $\mathcal{T}=\mathcal{B} \gets \mathcal{B}_r \cup \mathcal{B}_p;$ \\
    $\mathcal{S} \gets \mathcal{S}_r \cup \mathcal{S}_p$, $\mathcal{C} \gets \mathcal{C}_r \cup \mathcal{C}_p$\;
    \While{$\mathcal{T}\neq \varnothing$}{
         $m\gets argmax{ } \mathcal{S}$\;
         $\mathcal{M}\gets b_m$\;
         $\mathcal{T}\gets \mathcal{T} - \mathcal{M}$\;
         $\mathcal{S}\gets \mathcal{S}f(IoU(\mathcal{M},\mathcal{T}))$\;
        \If{$\mathcal{M} \in \mathcal{B}_r$}{
            \lIf{$\mathcal{B}_p \neq \varnothing$}{$iou \gets \max IoU(\mathcal{M},\mathcal{B}_p)$}
            \uIf(\tcc*[f]{Case1}){$iou \geq t_2$}{
                 $idx \gets IoU(\mathcal{M},\mathcal{B}) \geq t_2$\;
            }
            \uElseIf(\tcc*[f]{Case2}){$iou \geq t_1$}{
                 $idx \gets IoU(\mathcal{M},\mathcal{B}) \geq t_1$\;
            }
            \ElseIf(\tcc*[f]{Case3}){$iou < t_1\mathcal || {B}_p = \varnothing$}{$idx \gets IoU(\mathcal{M},\mathcal{B}_r) \geq t_1$\;}
        }
        \lElseIf{$\mathcal{M} \in \mathcal{B}_p$}{repeat algorithm above}
         map $idx$ to $\mathcal{B}$\;
         $\mathcal{M} \gets \mathcal{B}[idx]/\mathcal{C}[idx]/sum(1/\mathcal{C}[idx])$\;
         $\mathcal{D}\gets \mathcal{D} \cup \mathcal{M}$\;
    }
    \Return $\mathcal{D},\mathcal{S}$\;
\end{algorithm}

According to the definition above, we propose the extended softer-NMS in the Algorithm.\ref{NMS} to replace the joint NMS, titled uncertainty-aware multi-source NMS. To adapt the confidence attenuation strategy in softer-NMS, we need to merge and rerank the multi-modal predictions $A,B$ first. Because the predicted values from different modalities have a consistent range of values both in terms of box attributes and uncertainty, the fusion of predicted values becomes simpler. 

The algorithm first mixes all predictions in one pool to pick up a high-confidence box sequentially. Confidence smoothing\cite{Bodla2017SoftNMSI} will be applied for all these boxes and variable methods can be chosen. For boxes from each single modality, we calculate the IoU with boxes from other modal detections and generate three cases. The logic gate aims to filter out those that are mutually verified. And others will be mixed in voting progress like normal NMS. For Case1, we focus on the highly similar boxes from all modal data. For Case2, we apply general softer-NMS to fuse boxes adaptively. However, in Case3, boxes from another modality are severely off position or even empty, which is more likely to drop the localization accuracy. We ignore them to avoid potential influence from them. The iteration is conducted according to the confidence scores, and will not affect the box adjustment due to its disorderliness. Uncertainty scores are applied in the following step. This fusion method can easily be extended to more modes or more sensors with little need to change the algorithm.

\section{Experiment}
\subsection{Dataset}
\subsubsection{KITTI} We select the KITTI 2D object detection dataset\cite{Geiger2012AreWR} for problem investigation and model valuation, including 7481 pairs of camera images and LiDAR point clouds. It provides over 80K 2D boxes annotation with seven different classes: car, van, pedestrian, tram, van, truck, and cyclist. To avoid the influence of imbalance data distribution(we focus on multi-modal robustness), we combine them into three classes: car, pedestrian, and cyclist. In our experiment, the dataset is randomly split into train, validation, and test set by the ratio of 6:2:2. 

\subsubsection{Projection} Before training, we project the 3D point clouds onto the image plane to be 2D Depth images. Given a point $P_v=(x_v,y_v,z_v)^T$, we calculate:
\begin{equation}
P_v'=K_v[R_v|T_v]P_v
\end{equation}
where $K_v,R_v,T_v$ refer to the camera calibration matrix, rotation matrix, and translation matrix. The projected front-view point cloud depth map will be cropped to the same size as RGB images. We set the size as $128\times512$ in our experiment. Afterward, the value of both the depth map and the RGB images are normalized to [0,1] interval.

\subsubsection{Noise Simulation}
To achieve the upper bound of noise inference in Expr.\ref{definition}, we leverage the noise benchmark\cite{Michaelis2019BenchmarkingRI} and select sufficient noise levels to modify KITTI data. The benchmark contains 15 corruptions on 5 severity levels, including gaussian noise, blur, extreme weather, etc. Because some types of noise cannot be applied to the point clouds like weather aspects, we chose three representative interference methods: Gaussian noise, motion blur, and frost noise, with five-level noise intensity.

\begin{figure}[t!]
    \centering
    \includegraphics[width=1\linewidth]{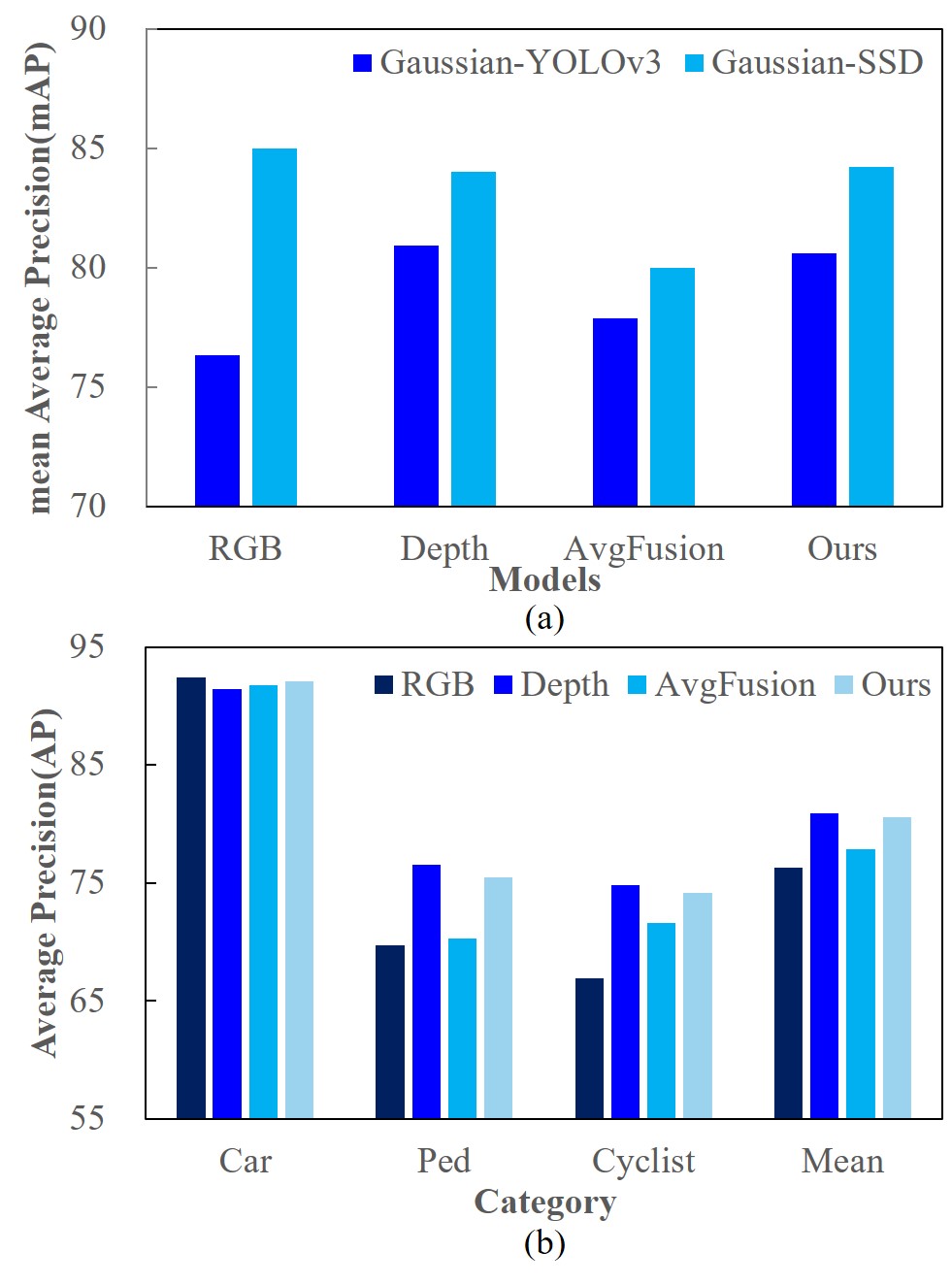}
    \caption{Performance on normal KITTI dataset. (a) presents the comparison based on Gaussian-YOLOv3\cite{Choi2019GaussianYA} and SSD\cite{Liu2016SSDSS}. (b) presents the details on three categories of Gaussian-YOLOv3.}
    \label{normal}
\end{figure}

\subsection{Implementation}
With the proposed multi-source NMS, we replace the joint NMS in the fusion model Fig.\ref{arch} with it. Our fusion strategy can be easily changed to be based on other types of NMS, or even adapted to NMS-free models. For fusion models, we first train the pipelines separately to provide stable convergence for fusion. Then we concatenate two pipelines with joint NMS or our proposed method for joint optimization. We set the IoU threshold in NMS as 0.45 for the two single-modal models, as set the two thresholds as 0.45 and 0.7 in our proposed models. 

We have also implemented our method with SSD\cite{Liu2016SSDSS} in the experiment, which can achieve even better performance for reference. For comparison, we use average box selection(AvgFusion) as the default joint NMS method. Half of the boxes from two pipelines will be randomly dropped, and processed in one NMS step. In all experiments, we set the batch size as 16 and the initial learning rate as 0.0001.The experiment is conducted on NVIDIA GTX 1080Ti with CUDA 8.0 and cuDNN v7.

\begin{table*}[]
\centering
\addtolength{\leftskip} {-2cm}
\addtolength{\rightskip}{-2cm}
\renewcommand\arraystretch{1.5}
\caption{AP for RGB/Depth single-modal models under level 1-5 noisy-data.}
\begin{tabular}{c|c|c|p{0.5cm}p{0.5cm}p{0.5cm}p{0.5cm}p{0.5cm}|p{0.5cm}p{0.5cm}p{0.5cm}p{0.5cm}p{0.5cm}|p{0.5cm}p{0.5cm}p{0.5cm}p{0.5cm}p{0.5cm}}
\hline
\multicolumn{1}{l|}{}          & \multicolumn{2}{c|}{Noise Type} & \multicolumn{5}{c|}{Gaussian Noise}  & \multicolumn{5}{c|}{Motion Blur}      & \multicolumn{5}{c}{Frost Noise}      \\ \hline
\textbf{Modality}               & \textbf{Category}               & Clean                 & L1     & L2     & L3     & L4     & L5    & L1     & L2     & L3     & L4     & L5     & L1     & L2     & L3     & L4     & L5     \\ \hline
\multirow{4}{*}{\textbf{RGB}}   
& \textbf{Car}                    & 92.39                 & 82.13 & 66.31 & 36.81 & 10.04 & 0.39 & 87.68 & 79.01 & 57.36 & 29.83 & 16.32 & 66.19 & 39.30 & 25.34 & 23.17 & 16.38 \\
& \textbf{Ped.}                    & 69.71                 & 41.83 & 21.71 & 6.68  & 0.76  & 0.00 & 40.04 & 19.79 & 7.13  & 0.97  & 0.21  & 21.75 & 4.86  & 1.18  & 1.55  & 0.59  \\ 
& \textbf{Cyclist}                & 66.92                 & 46.70 & 29.85 & 13.45 & 3.49  & 0.47 & 54.24 & 40.84 & 20.43 & 6.64  & 2.37  & 32.83 & 15.88 & 9.80  & 8.46  & 4.72  \\ 
& \textbf{mAP}                    & 76.34                 & 56.89 & 39.29 & 18.98 & 4.76  & 0.28 & 60.66 & 46.55 & 28.31 & 12.48 & 6.30  & 40.26 & 20.01 & 12.11 & 11.06 & 7.23  \\ \hline
\multirow{4}{*}{
\textbf{Depth}} 
& \textbf{Car}                    & 91.44                 & 83.68 & 70.02 & 43.84 & 18.98 & 2.38 & 87.23 & 84.40 & 76.80 & 63.74 & 53.25 & 59.11 & 36.51 & 25.54 & 26.09 & 20.84 \\
& \textbf{Ped.}                    & 76.53                 & 60.83 & 43.27 & 21.56 & 5.88  & 0.66 & 68.76 & 61.37 & 41.87 & 24.60 & 13.63 & 42.87 & 28.67 & 21.00 & 23.06 & 19.51 \\
& \textbf{Cyclist}                & 74.85                 & 64.71 & 53.84 & 36.32 & 18.43 & 2.04 & 67.03 & 60.81 & 48.68 & 36.76 & 27.99 & 53.24 & 44.58 & 37.50 & 37.94 & 34.24 \\
& \textbf{mAP}                    & 80.94                 & 69.74 & 55.71 & 33.90 & 14.43 & 1.69 & 74.34 & 68.86 & 55.78 & 41.70 & 31.62 & 51.74 & 36.59 & 28.01 & 29.03 & 24.86 \\ \hline
\end{tabular}
\label{singlemodal}
\end{table*}

\begin{table*}[]
\centering
\addtolength{\leftskip} {-2cm}
\addtolength{\rightskip}{-2cm}
\renewcommand\arraystretch{1.5}
\caption{AP for our fusion model under level 1-5 noisy-data. NR-D/R-ND indicate one modal data is with noise.}
\begin{tabular}{c|c|c|p{0.5cm}p{0.5cm}p{0.5cm}p{0.5cm}p{0.5cm}|p{0.5cm}p{0.5cm}p{0.5cm}p{0.5cm}p{0.5cm}|p{0.5cm}p{0.5cm}p{0.5cm}p{0.5cm}p{0.5cm}}
\hline
\multicolumn{1}{l|}{}          & \multicolumn{2}{c|}{Noise Type} & \multicolumn{5}{c|}{Gaussian Noise}  & \multicolumn{5}{c|}{Motion Blur}      & \multicolumn{5}{c}{Frost Noise}      \\ \hline
\textbf{Modality}               & \textbf{Category}               & Clean                 & L1     & L2     & L3     & L4     & L5    & L1     & L2     & L3     & L4     & L5     & L1     & L2     & L3     & L4     & L5     \\ \hline
\multirow{4}{*}{\textbf{NR-D}} 
& \textbf{Car}         & 92.10    & 90.36 & 88.97 & 87.75 & 87.78 & 88.16 & 90.91 & 89.11 & 87.14 & 86.54 & 87.38 & 88.62 & 88.06 & 87.85 & 87.76 & 87.72 \\ 
& \textbf{Ped.}        & 75.48    & 70.52 & 65.64 & 69.40 & 72.03 & 72.77 & 69.16 & 69.68 & 72.30 & 72.48 & 72.68 & 67.24 & 68.43 & 71.38 & 71.71 & 72.31 \\
& \textbf{Cyclist}     & 74.17    & 69.34 & 68.81 & 68.59 & 69.77 & 70.62 & 68.90 & 67.38 & 67.21 & 69.07 & 69.53 & 68.39 & 68.95 & 69.31 & 69.95 & 70.50 \\
& \textbf{mAP}         & 80.58    & 76.74 & 74.47 & 75.25 & 76.53 & 77.18 & 76.32 & 75.39 & 75.55 & 76.03 & 76.53 & 74.75 & 75.14 & 76.18 & 76.47 & 76.84 \\ \hline
\multirow{4}{*}{\textbf{R-ND}} 
& \textbf{Car}         & 92.10    & 91.40 & 90.60 & 90.62 & 90.73 & 90.87 & 91.97 & 91.44 & 89.79 & 88.17 & 87.40 & 90.31 & 90.41 & 90.62 & 90.72 & 90.70 \\
& \textbf{Ped.}        & 75.48    & 74.03 & 73.11 & 68.80 & 65.07 & 63.81 & 73.25 & 69.91 & 65.29 & 62.59 & 60.07 & 72.42 & 68.23 & 66.26 & 66.85 & 67.04 \\
& \textbf{Cyclist}     & 74.17    & 72.70 & 70.91 & 68.40 & 67.19 & 66.95 & 71.79 & 69.73 & 66.81 & 63.19 & 62.38 & 69.10 & 68.19 & 67.76 & 68.16 & 67.04 \\
& \textbf{mAP}         & 80.58    & 79.38 & 78.21 & 75.94 & 74.33 & 73.88 & 79.00 & 77.03 & 73.96 & 71.32 & 69.95 & 77.28 & 75.61 & 74.88 & 75.24 & 74.92 \\ \hline

\end{tabular}
\label{multimodal}
\end{table*}

\begin{table*}[]
\centering
\addtolength{\leftskip} {-2cm}
\addtolength{\rightskip}{-2cm}
\renewcommand\arraystretch{1.5}
\caption{Comparison between fusion models with/without data selection. The experiments were conducted under level 1-5 noisy-data. NR-ND indicates two modal data are noised.}
\begin{tabular}{c|c|c|p{0.5cm}p{0.5cm}p{0.5cm}p{0.5cm}p{0.5cm}|p{0.5cm}p{0.5cm}p{0.5cm}p{0.5cm}p{0.5cm}|p{0.5cm}p{0.5cm}p{0.5cm}p{0.5cm}p{0.5cm}}
\hline
\multicolumn{1}{l|}{}          & \multicolumn{2}{c|}{Noise Type} & \multicolumn{5}{c|}{Gaussian Noise}  & \multicolumn{5}{c|}{Motion Blur}      & \multicolumn{5}{c}{Frost Noise}      \\ \hline
\textbf{Modality}               & \textbf{Category}               & Clean                 & L1     & L2     & L3     & L4     & L5    & L1     & L2     & L3     & L4     & L5     & L1     & L2     & L3     & L4     & L5     \\ \hline
\multirow{4}{*}{\makecell{\textbf{NR-ND} \\ \textbf{w/o selection}}}
& \textbf{car}         & 91.92 & 87.41 & 80.04 & 66.12 & 53.21 & 46.65 & 89.69 & 86.81 & 79.50 & 69.35 & 63.35 & 77.28 & 64.91 & 58.68 & 58.27 & 55.26 \\
 & \textbf{ped}     & 73.12 & 62.23 & 52.81 & 43.62 & 38.22 & 36.73 & 63.76 & 56.85 & 48.81 & 42.95 & 40.02 & 52.72 & 44.94 & 42.11 & 42.71 & 41.59 \\
 & \textbf{cyclist} & 70.89 & 63.30 & 56.37 & 47.89 & 40.92 & 36.07 & 65.76 & 60.86 & 52.72 & 46.29 & 43.03 & 56.96 & 50.56 & 47.27 & 47.04 & 45.18 \\
 & \textbf{mAP}     & 78.64 & 70.98 & 63.07 & 52.54 & 44.12 & 39.81 & 73.07 & 68.17 & 60.34 & 52.87 & 48.80 & 62.32 & 53.47 & 49.35 & 49.34 & 47.34 \\ \hline
\multirow{4}{*}{\makecell{\textbf{NR-ND}\\\textbf{w/ selection}}}
& \textbf{Car}        & 92.10 & 90.88 & 89.79 & 89.19 & 89.26 & 89.52 & 91.44 & 90.28 & 88.47 & 87.36 & 87.39 & 89.47 & 89.24 & 89.24 & 89.24 & 89.21 \\
 & \textbf{ped}     & 75.48 & 72.28 & 69.38 & 69.10 & 68.55 & 68.29 & 71.21 & 69.80 & 68.80 & 67.54 & 66.38 & 69.83 & 68.33 & 68.82 & 69.28 & 69.68 \\
 & \textbf{cyclist} & 74.17 & 71.02 & 69.86 & 68.50 & 68.48 & 68.79 & 70.35 & 68.56 & 67.01 & 66.13 & 65.96 & 68.75 & 68.57 & 68.54 & 69.06 & 68.77 \\
 & \textbf{mAP}     & 80.58 & 78.06 & 76.34 & 75.60 & 75.43 & 75.53 & 77.66 & 76.21 & 74.76 & 73.68 & 73.24 & 76.02 & 75.38 & 75.53 & 75.86 & 75.88 \\ \hline

\end{tabular}
\label{comparison}
\end{table*}

\subsection{Basic Results}
Preliminary experimental results on the original KITTI dataset are shown in the results are shown in Fig.\ref{normal}. In Fig.\ref{normal}(a), we compared two implementation versions based on Gaussian-YOLOv3\cite{Choi2019GaussianYA} and SSD\cite{Liu2016SSDSS}, which has been revised to Gaussian-SSD. The results showed that SSD achieved better performance in all models. That means our approach can generalize to other detection models and achieve better performance by incorporating a better baseline. 

However, in the subsequent experiments, we choose to use Gaussian-YOLOv3 model to highlight the effect of the fusion algorithm. We present the comparison on three main categories in Fig.\ref{normal}(b). Both AvgFusion and our model achieve the sub-optimal performance between RGB and depth models due to the lack of prior information in box selection/fusion. But our model has higher accuracy, and approaches the optimal modal performance on all objects, which demonstrates the benefits of our method on clean data, and the accuracy of our informative data selection that reflects the quality of fusion object boxes.

\section{Overall Analysis}
We have further investigated the model's performance with noisy KITTI data for a comprehensive understanding of the robustness in multi-modal fusion. First, we evaluate the performance of single-modal models with noisy-data to validate the simulated noise in Table.\ref{singlemodal}. Then, we conduct a similar test for our proposed fusion model in Table.\ref{multimodal}.  NR-D means fusion with noisy RGB data, while R-ND means fusion with noisy depth data. We finally provide supplement conclusions based on the results. In all experiments, models are trained with clean data and tested with noisy data.

\subsection{Detection Model Degradation under Noise}
The experiments show that the single-modal model is severely affected by the noisy data, and the effect increases with the degree of noise disturbance. The experimental results are shown in Table.\ref{singlemodal}. The numbers under the noise name indicate the inference level. We list the mAP of the proposed model under different noise settings. Among them, the prediction accuracy of the RGB model shows a nonlinear decrease for the three types of noise, with mAP decreasing from 76.34 to 0.28(Gaussian), 6.30(Motion), and 7.23(Frost), respectively. Compared with the detection results of clean data, the accuracy of all 15 experiments decreases. Similarly, the prediction accuracy of the Depth model also decreases nonlinearly with increasing noise intensity, with mAP decreasing from 76.34 to 1.69(Gaussian), 31.62(Motion), and 24.86(Frost), respectively. The accuracy decreases in 14/15 experiments. 

In addition, the sensitivity of both modal data to these noises is consistent. The severity of the noise impact, from greatest to least, is
\begin{equation}
    Gaussian\ Noise>Frost\ Noise>Motion\ Blur
\end{equation}
In most cases, for the same class and interference level of noise, the Depth model has higher detection accuracy. In all 15 sets of experiments, the Depth model outperformed the RGB model in terms of accuracy, which is the same as the comparison results on clean data. The average accuracy increments are, 7.74+ for Car, 20.57+ for Pedestrian, 22.26+ for Cyclist, and 16.86+ for mAP.

\subsection{Informative Data Selection is Generalized to Noise}
Through the Fig.\ref{normal}(b) we learn that on clean data, our model obtains sub-optimal results close to the highest single-modal accuracy. Further, we tested the performance of the fusion model in the case subjected to single-modal data noise, as shown in the Table.\ref{multimodal}.

The results show that the degradation of detection accuracy is smaller when our proposed RGB-Depth fusion model is affected by the noise of only one modality. When the fusion model receives noisy RGB and Depth data(NR-D), only 5/15 experiments show accuracy degradation. Its average accuracy changes for the three RGB noisy data are -3.40 (Gaussian), -4.05 (Motion), and -3.74 (Frost). mAP decreases from 80.58 to 77.18, 76.53, and 76.84, respectively. it can be seen that the noise for the RGB data has very little effect on our proposed fusion model. The effect of noise on our proposed fusion model is very small. In addition, the detection error of Car is positively correlated with the noise intensity, but the other categories are not. This leads to a slight increase of mAP from the lowest point during the increasing noise intensity (in 10/15 experiments). This issue needs to be further explored in the subsequent work.

Such a situation also exists for the noisy Depth data. When the fusion model receives RGB and noisy Depth data(R-ND), the average accuracy decreases are: -6.70(Gaussian), -10.63(Motion), and -5.66(Frost), respectively, although there is a decrease in accuracy in 14/15 experiments. mAP decreases from 80.58 to 73.88, 69.95, and 74.92. However, there is no increase in mAP with increasing noise intensity.

Furthermore, we compare the results on the same models with/without our informative data selection in fusion in Table.\ref{comparison}. It shows that even simple fusion can improve performance compared with single-modal models, but selective fusion can exceed it significantly in noisy data.

Overall, our proposed fusion model is robust to single modal data noise and there is no substantial change in detection accuracy. Next, we will analyze the gain of the fusion model on the single-modal models.


\subsection{Further Investigation on Noisy-data Fusion}
We further compare the fusion model with the unimodal model. Although the fusion model outperforms the unimodal model on noisy data, this is not incremental, as it may be degrading for the unimodal model with clean data. For example, the fusion model that received noisy data outperformed the RGB model in all tests (average mAP increase of 51.61), but not as well as the Depth model and the fusion model with clean data: NRGB$<$NR-D$<$Depth$<$Fusion.
\begin{equation}
    NRGB<NR-D<Depth<Fusion
\end{equation}
The fusion model that received noisy data was better than the Depth model in all tests (average mAP increase of 34.20), but not as good as the RGB model and the fusion model with clean data. When there is less noise, 
\begin{equation}
    NDepth<RGB<R-ND<Fusion
\end{equation}
when there is more noise, 
\begin{equation}
    NDepth<R-ND<RGB<Fusion
\end{equation}
The results suggest that models with noisy data improve when they fuse clean data, and conversely, models with clean data may deteriorate when they fuse noisy data. However, there is also a smaller probability of an increase in accuracy in specific experiments, although that is difficult to explain now.

In addition, the sensitivity of fusion to RGB noise is: 
\begin{equation}
    Motion\ Blur>Frost>Gaussian
\end{equation}
and to Depth noise is: 
\begin{equation}
    Motion\ Blur>Gaussian>Frost
\end{equation}
, but overall they both have a relatively small effect. The average accuracy decreases are: -3.73(NR-D) and -7.66(R-ND). For specific detection targets, RGB noise in multi-modal data has a greater effect on Car, and NR-D has lower detection accuracy than R-ND in all 15 experiments, with an average of 2.11 lower. In contrast, Depth noise has a greater effect on Pedestrian and Cyclist, and R-ND outperforms the NR-D model in most experiments. RGB noise has more influence when noise-level$<$3, and Depth noise has more influence when noise intensity rises. 

These findings illustrate the complexity of the performance of the fusion model in the face of noisy data. For different targets, recognition tasks, data modality combinations, and noise types, fusion models may exhibit different performances. They do not show a simple linear or nonlinear change for increasing noise intensity, and may even show an increase in accuracy, which requires to be further investigated.

\subsection{Rethink Our Model and Experiments}
The ablation study has been included in the comparative analysis of the single-modal and multi-modal models and are therefore not listed separately. In the following, We rethink our work in terms of data noise, data selection, and fusion models.

\subsubsection{Multi-modal Noise}
We selected three general and suitable noises from the image noise benchmark to add to the RGB and Depth data because it is difficult to capture or simulate the corresponding noise data in real scenes. However, due to the gap between data modalities, such an approach still has a large problem to fit the real error and will lead to potential bias in the experimental data. However, for the robustness of the fusion model discussed in this paper, we mainly focus on adding noise interference of sufficient strength, so that such an error may be negligible.

\subsubsection{Data Selection Accuracy}
\begin{figure}[t!]
    \centering
    \includegraphics[width=1\linewidth]{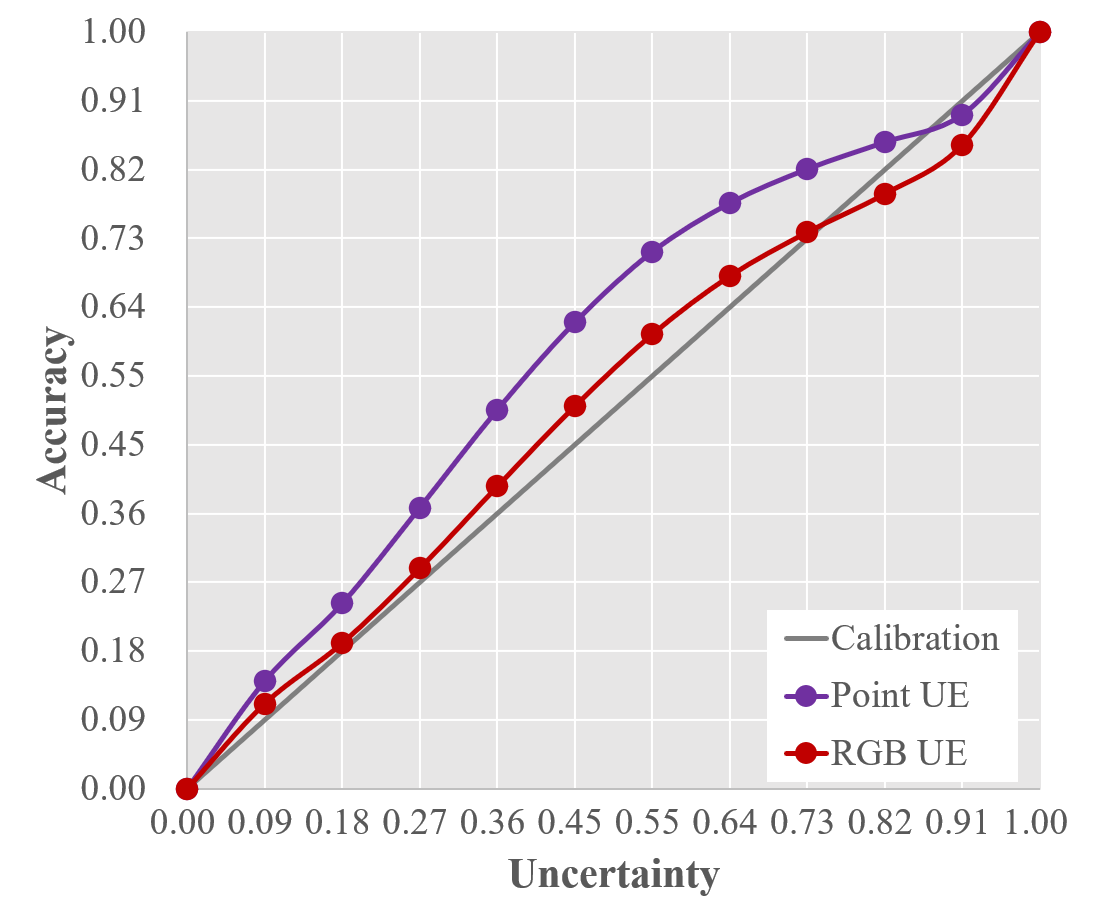}
    \caption{Uncertainty calibration for the proposed model.}
    \label{calib}
\end{figure}

In our setting, the selection accuracy mainly relies on uncertainty estimation. Uncertainty estimation is a prerequisite of the method proposed in this paper, and thus the accuracy of the estimation determines the interpretability of the fusion model. Therefore, we refer to the ECE method for uncertainty calibration\cite{Kuleshov2018AccurateUF}, as shown in Fig.\ref{calib} We model the uncertainty as a Gaussian distribution of the bounding box estimation parameters, so the magnitude of the variance corresponds to the potential range of values taken. The error between the true and estimated values can be portrayed by the distance between the curve and the $y=x$ line. The RGB uncertainty estimates are more accurate, while Depth estimates have deviation, which can also lead to potential model bias.

\subsubsection{Fusion Network Design}
We have further implemented the feature fusion model with typical methods\cite{Zou2021EfficientUP}. However it is easy to have a dependence on a single modality, i.e., there is a significant primary and secondary relationship. The test results are shown in the Table.\ref{featurefusion}. We apply weighted feature addition at seven layers in the two-pipeline network. The $2\times7$ weight parameters will be optimized during training. When trained without any limitation, the network tends to assign more weights for the Depth branch, which will seriously affect the robustness of the model. Then we use imbalance data dropout(10\% for RGB and 15\% for Depth) and weight limitation(upper to 0.7 in the first five stages) to adjust the fusion weight in training. The adjustment slightly improved the performance. But it is far not enough, and adaptive balance fusion remains an open problem.

\begin{table}[h!]
\centering
\renewcommand\arraystretch{1.5}
\caption{Modality bias in feature fusion. L1-L7 indicate seven fusion layers from front to back in the Darknet in YOLOv3. NR and ND represent Noisy-RGB and Noisy-Depth. w/ and w/o indicate whether we limit the fusion weights in training.}
\begin{tabular}{|p{0.4cm}|c|p{0.3cm}|p{0.3cm}|p{0.3cm}|p{0.3cm}|p{0.3cm}|p{0.3cm}|p{0.3cm}|p{0.4cm}|p{0.4cm}|}
\hline
\textbf{Exp.}       & \textbf{Modal} & \textbf{L1} & \textbf{L2} & \textbf{L3} & \textbf{L4} & \textbf{L5} & \textbf{L6} & \textbf{L7} & \textbf{NR}           & \textbf{ND}           \\ \hline
\multirow{2}{*}{w/o} & RGB   & 0.3        & 0.6        & 0.7        & 0.7        & 0.5        & 0.5        & 0.1        & \multirow{2}{*}{83.2} & \multirow{2}{*}{19.8} \\ \cline{2-9}
                     & Depth & 0.7        & 0.4        & 0.7        & 0.7        & 0.7        & 0.9        & 0.7        &                       &                       \\ \hline
\multirow{2}{*}{w/}  & RGB   & 0.6        & 0.5        & 0.8        & 0.8        & 0.7        & 0.4        & 0.3        & \multirow{2}{*}{77.6} & \multirow{2}{*}{39.5} \\ \cline{2-9}
                     & Depth & 0.3        & 0.1        & 0.2        & 0.1        & 0.2        & 0.8        & 0.7        &                       &                       \\ \hline
\end{tabular}
\label{featurefusion}
\end{table}

\section{Conclusion}
In this paper, we explore the robustness of single-modal and multi-modal models under noisy data, and claim that informative data selection is a simple but practical method in data fusion and anti-noise vision tasks. We present the conclusions on the relationship between data information and the performance of deep learning models, especially the performance of multimodal fusion models. It is pointed out that the information-driven selection strategy can improve the robustness of the fusion model. In addition, uncertainty can be used as an effective method to approximate the amount of data information. We analyze and reveal the robustness of multi-modal models on dirty data by comparing the accuracy decay of different models through experiments on the KITTI dataset. The experimental results show that the impact of noisy data on single-modal and multi-modal models is complex and involves many aspects like recognition target, recognition task, data modality combinations, noise type, etc. We also propose a novel multi-modal fused 2D target detection model. It performs selective fusion of bounding boxes generated by multiple independent sub-models based on informative data selection with uncertainty and can be applied to different modal data. Experimental results show that our fusion model can exhibit high detection accuracy when a clean data modality is present under severe noise interference. In addition, the robustness of the fusion model may show different results for different fusion structures, such as feature fusion. The robustness problem of multi-modal fusion models still needs to be studied in depth, and we will also focus on other tasks such as semantic segmentation and 3D target detection in our subsequent research to further investigate the phenomena observed in this paper.

As for another trend in research, although there is no literature specifically addressing this issue, it can be inferred from the latest research trends that a large model based on transformer can hopefully handle this issue well after training with enough data. However, this article hopes to present a simple and effective solution based on a non-parametric perspective. Compared with fitting the rich and varied real world, our method of directly measuring the information content of data provides researchers with a simple approach.


%





\ifCLASSOPTIONcaptionsoff
  \newpage
\fi



%



\bibliographystyle{IEEEtran}
\bibliography{IEEEabrv,mybib}

\begin{thebibliography}{10}
\providecommand{\url}[1]{#1}
\csname url@samestyle\endcsname
\providecommand{\newblock}{\relax}
\providecommand{\bibinfo}[2]{#2}
\providecommand{\BIBentrySTDinterwordspacing}{\spaceskip=0pt\relax}
\providecommand{\BIBentryALTinterwordstretchfactor}{4}
\providecommand{\BIBentryALTinterwordspacing}{\spaceskip=\fontdimen2\font plus
\BIBentryALTinterwordstretchfactor\fontdimen3\font minus
  \fontdimen4\font\relax}
\providecommand{\BIBforeignlanguage}[2]{{%
\expandafter\ifx\csname l@#1\endcsname\relax
\typeout{** WARNING: IEEEtran.bst: No hyphenation pattern has been}%
\typeout{** loaded for the language `#1'. Using the pattern for}%
\typeout{** the default language instead.}%
\else
\language=\csname l@#1\endcsname
\fi
#2}}
\providecommand{\BIBdecl}{\relax}
\BIBdecl

\bibitem{Feng2021DeepMO}
D.~Feng, C.~Haase-Schuetz, L.~Rosenbaum, H.~Hertlein, F.~Duffhauss,
  C.~Gl{\"a}ser, W.~Wiesbeck, and K.~C.~J. Dietmayer, ``Deep multi-modal object
  detection and semantic segmentation for autonomous driving: Datasets,
  methods, and challenges,'' \emph{IEEE Transactions on Intelligent
  Transportation Systems}, vol.~22, pp. 1341--1360, 2021.

\bibitem{Bijelic2018RobustnessAU}
M.~Bijelic, C.~Muench, W.~Ritter, Y.~Kalnishkan, and K.~C.~J. Dietmayer,
  ``Robustness against unknown noise for raw data fusing neural networks,''
  \emph{2018 21st International Conference on Intelligent Transportation
  Systems (ITSC)}, pp. 2177--2184, 2018.

\bibitem{Bijelic2020SeeingTF}
M.~Bijelic, T.~Gruber, F.~Mannan, F.~Kraus, W.~Ritter, K.~C.~J. Dietmayer, and
  F.~Heide, ``Seeing through fog without seeing fog: Deep multimodal sensor
  fusion in unseen adverse weather,'' \emph{2020 IEEE/CVF Conference on
  Computer Vision and Pattern Recognition (CVPR)}, pp. 11\,679--11\,689, 2020.

\bibitem{Hnewa2021ObjectDU}
M.~Hnewa and H.~Radha, ``Object detection under rainy conditions for autonomous
  vehicles: A review of state-of-the-art and emerging techniques,'' \emph{IEEE
  Signal Processing Magazine}, vol.~38, pp. 53--67, 2021.

\bibitem{Ma2021PointDropIO}
W.~Ma, J.~Chen, Q.~Du, and W.~Jia, ``Pointdrop: Improving object detection from
  sparse point clouds via adversarial data augmentation,'' \emph{2020 25th
  International Conference on Pattern Recognition (ICPR)}, pp.
  10\,004--10\,009, 2021.

\bibitem{Zhuang2021ACS}
F.~Zhuang, Z.~Qi, K.~Duan, D.~Xi, Y.~Zhu, H.~Zhu, H.~Xiong, and Q.~He, ``A
  comprehensive survey on transfer learning,'' \emph{Proceedings of the IEEE},
  vol. 109, pp. 43--76, 2021.

\bibitem{Yi2020AdaptiveWS}
S.~Yi, Z.~He, X.~Jing, Y.~Li, Y.~ming Cheung, and F.~Nie, ``Adaptive weighted
  sparse principal component analysis for robust unsupervised feature
  selection,'' \emph{IEEE Transactions on Neural Networks and Learning
  Systems}, vol.~31, pp. 2153--2163, 2020.

\bibitem{Yang2020LaNoisingAD}
T.~Yang, Y.~Li, Y.~Ruichek, and Z.~Yan, ``Lanoising: A data-driven approach for
  903nm tof lidar performance modeling under fog,'' \emph{2020 IEEE/RSJ
  International Conference on Intelligent Robots and Systems (IROS)}, pp.
  10\,084--10\,091, 2020.

\bibitem{Baltruaitis2019MultimodalML}
T.~Baltrusaitis, C.~Ahuja, and L.-P. Morency, ``Multimodal machine learning: A
  survey and taxonomy,'' \emph{IEEE Transactions on Pattern Analysis and
  Machine Intelligence}, vol.~41, pp. 423--443, 2019.

\bibitem{Geiger2012AreWR}
A.~Geiger, P.~Lenz, and R.~Urtasun, ``Are we ready for autonomous driving? the
  kitti vision benchmark suite,'' \emph{2012 IEEE Conference on Computer Vision
  and Pattern Recognition}, pp. 3354--3361, 2012.

\bibitem{Caesar2020nuScenesAM}
H.~Caesar, V.~Bankiti, A.~H. Lang, S.~Vora, V.~E. Liong, Q.~Xu, A.~Krishnan,
  Y.~Pan, G.~Baldan, and O.~Beijbom, ``nuscenes: A multimodal dataset for
  autonomous driving,'' \emph{2020 IEEE/CVF Conference on Computer Vision and
  Pattern Recognition (CVPR)}, pp. 11\,618--11\,628, 2020.

\bibitem{Sun2020ScalabilityIP}
P.~Sun, H.~Kretzschmar, X.~Dotiwalla, A.~Chouard, V.~Patnaik, P.~Tsui, J.~Guo,
  Y.~Zhou, Y.~Chai, B.~Caine, V.~Vasudevan, W.~Han, J.~Ngiam, H.~Zhao,
  A.~Timofeev, S.~M. Ettinger, M.~Krivokon, A.~Gao, A.~Joshi, Y.~Zhang,
  J.~Shlens, Z.~Chen, and D.~Anguelov, ``Scalability in perception for
  autonomous driving: Waymo open dataset,'' \emph{2020 IEEE/CVF Conference on
  Computer Vision and Pattern Recognition (CVPR)}, pp. 2443--2451, 2020.

\bibitem{Oh2017ObjectDA}
S.-I. Oh and H.-B. Kang, ``Object detection and classification by
  decision-level fusion for intelligent vehicle systems,'' \emph{Sensors
  (Basel, Switzerland)}, vol.~17, 2017.

\bibitem{Xu2018PointFusionDS}
D.~Xu, D.~Anguelov, and A.~Jain, ``Pointfusion: Deep sensor fusion for 3d
  bounding box estimation,'' \emph{2018 IEEE/CVF Conference on Computer Vision
  and Pattern Recognition}, pp. 244--253, 2018.

\bibitem{Liang2018DeepCF}
M.~Liang, B.~Yang, S.~Wang, and R.~Urtasun, ``Deep continuous fusion for
  multi-sensor 3d object detection,'' in \emph{ECCV}, 2018.

\bibitem{Qi2018FrustumPF}
C.~Qi, W.~Liu, C.~Wu, H.~Su, and L.~J. Guibas, ``Frustum pointnets for 3d
  object detection from rgb-d data,'' \emph{2018 IEEE/CVF Conference on
  Computer Vision and Pattern Recognition}, pp. 918--927, 2018.

\bibitem{Feng2020LeveragingUF}
D.~Feng, Y.~Cao, L.~Rosenbaum, F.~Timm, and K.~C.~J. Dietmayer, ``Leveraging
  uncertainties for deep multi-modal object detection in autonomous driving,''
  \emph{2020 IEEE Intelligent Vehicles Symposium (IV)}, pp. 877--884, 2020.

\bibitem{Zou2021ANM}
Z.~Zou, X.~Zhang, H.~Liu, Z.~Li, A.~Hussain, and J.~Li, ``A novel multimodal
  fusion network based on a joint coding model for lane line segmentation,''
  \emph{Information Fusion}, vol.~80, pp. 167--178, 2022.

\bibitem{Michaelis2019BenchmarkingRI}
C.~Michaelis, B.~Mitzkus, R.~Geirhos, E.~Rusak, O.~Bringmann, A.~S. Ecker,
  M.~Bethge, and W.~Brendel, ``Benchmarking robustness in object detection:
  Autonomous driving when winter is coming,'' \emph{ArXiv}, vol.
  abs/1907.07484, 2019.

\bibitem{Kendall2017WhatUD}
A.~Kendall and Y.~Gal, ``What uncertainties do we need in bayesian deep
  learning for computer vision?'' in \emph{NIPS}, 2017.

\bibitem{OforiOduro2020DataAU}
M.~Ofori-Oduro and M.~Amer, ``Data augmentation using artificial immune systems
  for noise-robust cnn models,'' \emph{2020 IEEE International Conference on
  Image Processing (ICIP)}, pp. 833--837, 2020.

\bibitem{Loh2019GettingTK}
Y.~P. Loh and C.~S. Chan, ``Getting to know low-light images with the
  exclusively dark dataset,'' \emph{Comput. Vis. Image Underst.}, vol. 178, pp.
  30--42, 2019.

\bibitem{Khodabandeh2019ARL}
M.~Khodabandeh, A.~Vahdat, M.~Ranjbar, and W.~G. Macready, ``A robust learning
  approach to domain adaptive object detection,'' \emph{2019 IEEE/CVF
  International Conference on Computer Vision (ICCV)}, pp. 480--490, 2019.

\bibitem{Tian2020UNOUN}
J.~Tian, W.~Cheung, N.~Glaser, Y.-C. Liu, and Z.~Kira, ``Uno: Uncertainty-aware
  noisy-or multimodal fusion for unanticipated input degradation,'' \emph{2020
  IEEE International Conference on Robotics and Automation (ICRA)}, pp.
  5716--5723, 2020.

\bibitem{Qi2017PointNetDL}
C.~Qi, H.~Su, K.~Mo, and L.~J. Guibas, ``Pointnet: Deep learning on point sets
  for 3d classification and segmentation,'' \emph{2017 IEEE Conference on
  Computer Vision and Pattern Recognition (CVPR)}, pp. 77--85, 2017.

\bibitem{Vora2020PointPaintingSF}
S.~Vora, A.~H. Lang, B.~Helou, and O.~Beijbom, ``Pointpainting: Sequential
  fusion for 3d object detection,'' \emph{2020 IEEE/CVF Conference on Computer
  Vision and Pattern Recognition (CVPR)}, pp. 4603--4611, 2020.

\bibitem{Valada2017AdapNetAS}
A.~Valada, J.~Vertens, A.~Dhall, and W.~Burgard, ``Adapnet: Adaptive semantic
  segmentation in adverse environmental conditions,'' \emph{2017 IEEE
  International Conference on Robotics and Automation (ICRA)}, pp. 4644--4651,
  2017.

\bibitem{Valada2019SelfSupervisedMA}
A.~Valada, R.~Mohan, and W.~Burgard, ``Self-supervised model adaptation for
  multimodal semantic segmentation,'' \emph{International Journal of Computer
  Vision}, vol. 128, pp. 1239--1285, 2019.

\bibitem{Mees2016ChoosingSA}
O.~Mees, A.~Eitel, and W.~Burgard, ``Choosing smartly: Adaptive multimodal
  fusion for object detection in changing environments,'' \emph{2016 IEEE/RSJ
  International Conference on Intelligent Robots and Systems (IROS)}, pp.
  151--156, 2016.

\bibitem{Zhang2021ProgressiveMC}
W.~Zhang, D.~Xu, J.~Zhang, and W.~Ouyang, ``Progressive modality cooperation
  for multi-modality domain adaptation,'' \emph{IEEE Transactions on Image
  Processing}, vol.~30, pp. 3293--3306, 2021.

\bibitem{Zhao2021AdaptiveCM}
S.~Zhao, M.~Gong, H.~Fu, and D.~Tao, ``Adaptive context-aware multi-modal
  network for depth completion,'' \emph{IEEE Transactions on Image Processing},
  vol.~30, pp. 5264--5276, 2021.

\bibitem{Kim2018RobustDM}
J.~Kim, J.~Koh, Y.~Kim, J.~Choi, Y.~Hwang, and J.~W. Choi, ``Robust deep
  multi-modal learning based on gated information fusion network,'' in
  \emph{ACCV}, 2018.

\bibitem{Liu2016SSDSS}
W.~Liu, D.~Anguelov, D.~Erhan, C.~Szegedy, S.~E. Reed, C.-Y. Fu, and A.~C.
  Berg, ``Ssd: Single shot multibox detector,'' in \emph{ECCV}, 2016.

\bibitem{Kim2019OnSS}
T.~Kim and J.~Ghosh, ``On single source robustness in deep fusion models,'' in
  \emph{NeurIPS}, 2019.

\bibitem{Lee2021DBFDB}
H.~Lee and H.~Kwon, ``Dbf: Dynamic belief fusion for combining multiple object
  detectors,'' \emph{IEEE Transactions on Pattern Analysis and Machine
  Intelligence}, vol.~43, pp. 1499--1514, 2021.

\bibitem{shannon2001mathematical}
C.~E. Shannon, ``A mathematical theory of communication,'' \emph{ACM SIGMOBILE
  mobile computing and communications review}, vol.~5, no.~1, pp. 3--55, 2001.

\bibitem{mackay2003information}
D.~J. MacKay, D.~J. Mac~Kay \emph{et~al.}, \emph{Information theory, inference
  and learning algorithms}.\hskip 1em plus 0.5em minus 0.4em\relax Cambridge
  university press, 2003.

\bibitem{tishby2000information}
N.~Tishby, F.~C. Pereira, and W.~Bialek, ``The information bottleneck method,''
  \emph{arXiv preprint physics/0004057}, 2000.

\bibitem{tishby2015deep}
N.~Tishby and N.~Zaslavsky, ``Deep learning and the information bottleneck
  principle,'' in \emph{2015 ieee information theory workshop (itw)}.\hskip 1em
  plus 0.5em minus 0.4em\relax IEEE, 2015, pp. 1--5.

\bibitem{belghazi2018mutual}
M.~I. Belghazi, A.~Baratin, S.~Rajeshwar, S.~Ozair, Y.~Bengio, A.~Courville,
  and D.~Hjelm, ``Mutual information neural estimation,'' in
  \emph{International conference on machine learning}.\hskip 1em plus 0.5em
  minus 0.4em\relax PMLR, 2018, pp. 531--540.

\bibitem{zou2020mimf}
Z.~Zou, L.~Zhao, X.~Zhang, Z.~Li, D.~Jin, and T.~Luo, ``Mimf: Mutual
  information-driven multimodal fusion,'' in \emph{International Conference on
  Cognitive Systems and Signal Processing}.\hskip 1em plus 0.5em minus
  0.4em\relax Springer, 2020, pp. 142--150.

\bibitem{Deng2019DomainAV}
W.-Y. Deng, A.~Lendasse, Y.~S. Ong, I.~W.-H. Tsang, L.~Chen, and Q.~Zheng,
  ``Domain adaption via feature selection on explicit feature map,'' \emph{IEEE
  Transactions on Neural Networks and Learning Systems}, vol.~30, pp.
  1180--1190, 2019.

\bibitem{xu2020u2fusion}
H.~Xu, J.~Ma, J.~Jiang, X.~Guo, and H.~Ling, ``U2fusion: A unified unsupervised
  image fusion network,'' \emph{IEEE Transactions on Pattern Analysis and
  Machine Intelligence}, vol.~44, no.~1, pp. 502--518, 2020.

\bibitem{Gal2016UncertaintyID}
Y.~Gal, ``Uncertainty in deep learning,'' 2016.

\bibitem{He2019BoundingBR}
Y.~He, C.~Zhu, J.~Wang, M.~Savvides, and X.~Zhang, ``Bounding box regression
  with uncertainty for accurate object detection,'' \emph{2019 IEEE/CVF
  Conference on Computer Vision and Pattern Recognition (CVPR)}, pp.
  2883--2892, 2019.

\bibitem{Choi2019GaussianYA}
J.~Choi, D.~Chun, H.~Kim, and H.-J. Lee, ``Gaussian yolov3: An accurate and
  fast object detector using localization uncertainty for autonomous driving,''
  \emph{2019 IEEE/CVF International Conference on Computer Vision (ICCV)}, pp.
  502--511, 2019.

\bibitem{Lee2020LocalizationUE}
Y.~Lee, J.~Hwang, H.-I. Kim, K.~Yun, and J.~Park, ``Localization uncertainty
  estimation for anchor-free object detection,'' \emph{ArXiv}, vol.
  abs/2006.15607, 2020.

\bibitem{Kowol2021YOdarUS}
K.~Kowol, M.~Rottmann, S.~Bracke, and H.~Gottschalk, ``Yodar: Uncertainty-based
  sensor fusion for vehicle detection with camera and radar sensors,'' in
  \emph{ICAART}, 2021.

\bibitem{Redmon2018YOLOv3AI}
J.~Redmon and A.~Farhadi, ``Yolov3: An incremental improvement,'' \emph{ArXiv},
  vol. abs/1804.02767, 2018.

\bibitem{Feng2018TowardsSA}
D.~Feng, L.~Rosenbaum, and K.~C.~J. Dietmayer, ``Towards safe autonomous
  driving: Capture uncertainty in the deep neural network for lidar 3d vehicle
  detection,'' \emph{2018 21st International Conference on Intelligent
  Transportation Systems (ITSC)}, pp. 3266--3273, 2018.

\bibitem{Feng2019LeveragingHA}
D.~Feng, L.~Rosenbaum, F.~Timm, and K.~C.~J. Dietmayer, ``Leveraging
  heteroscedastic aleatoric uncertainties for robust real-time lidar 3d object
  detection,'' \emph{2019 IEEE Intelligent Vehicles Symposium (IV)}, pp.
  1280--1287, 2019.

\bibitem{Russell2019MultivariateUI}
R.~L. Russell and C.~Reale, ``Multivariate uncertainty in deep learning,''
  \emph{IEEE Transactions on Neural Networks and Learning Systems}, vol.~33,
  pp. 7937--7943, 2022.

\bibitem{Bodla2017SoftNMSI}
N.~Bodla, B.~Singh, R.~Chellappa, and L.~S. Davis, ``Soft-nms — improving
  object detection with one line of code,'' \emph{2017 IEEE International
  Conference on Computer Vision (ICCV)}, pp. 5562--5570, 2017.

\bibitem{Kuleshov2018AccurateUF}
V.~Kuleshov, N.~Fenner, and S.~Ermon, ``Accurate uncertainties for deep
  learning using calibrated regression,'' \emph{ArXiv}, vol. abs/1807.00263,
  2018.

\bibitem{Zou2021EfficientUP}
Z.~Zou and Y.~Li, ``Efficient urban-scale point clouds segmentation with bev
  projection,'' \emph{ArXiv}, vol. abs/2109.09074, 2021.

\end{thebibliography}

%

\begin{IEEEbiography}[{\includegraphics[width=1in,height=1.25in,clip,keepaspectratio]{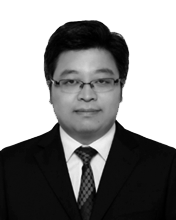}}]{Xinyu Zhang} was born in Huining, Gansu Province, and he received the B.E. degree from the School of Vehicle and Mobility at Tsinghua University, in 2001. He was a visiting scholar at University of Cambridge. He is currently a researcher with the School of Vehicle and Mobility, and the head of the Mengshi Intelligent Vehicle Team at Tsinghua University. He is the author of more than 30 SCI/EI articles. His research interests include intelligent driving and multimodal information fusion.
\end{IEEEbiography}

\begin{IEEEbiography}[{\includegraphics[width=1in,height=1.25in,clip,keepaspectratio]{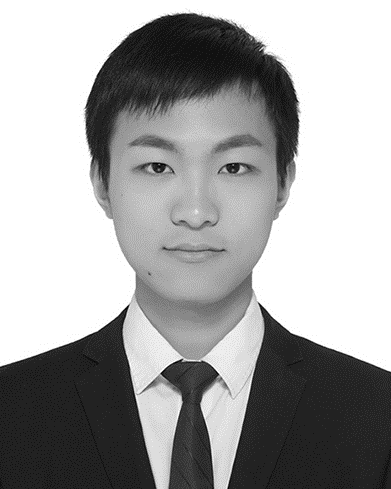}}]{Zhenhong Zou} was born in Zhanjiang, Guangdong Province, China in 1998. He received his B.S. degree in Information and Computation Science from Beihang University, in 2020. He was a visiting student with the Dept. of Mathematics at the University of California, Los Angeles, in 2019. After that, he has been a research assistant with the School of Vehicle and Mobility at Tsinghua University. His research interests include machine learning, multi-sensor fusion and computer vision.
\end{IEEEbiography}

\begin{IEEEbiography}[{\includegraphics[width=1in,height=1.25in,clip,keepaspectratio]{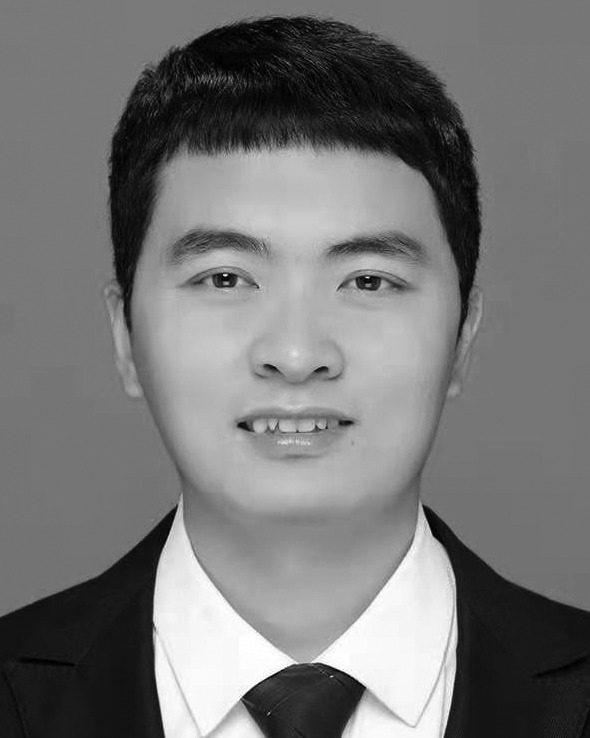}}]{Zhiwei Li} was born in Lvliang, Shanxi Province, and he received a Ph.D. degree from the China University of Mining \& Technology,Beijing, in 2020. Currently, he is a tutor of master students in Beijing University of Chemical Technology. In 2020, he studied as a postdoctoral fellow with Academician Jun Li at Tsinghua University. His main research interests include computer vision, intelligent perception and autonomous driving, and robot system architecture.
\end{IEEEbiography}

\begin{IEEEbiography}[{\includegraphics[width=1in,height=1.25in,clip,keepaspectratio]{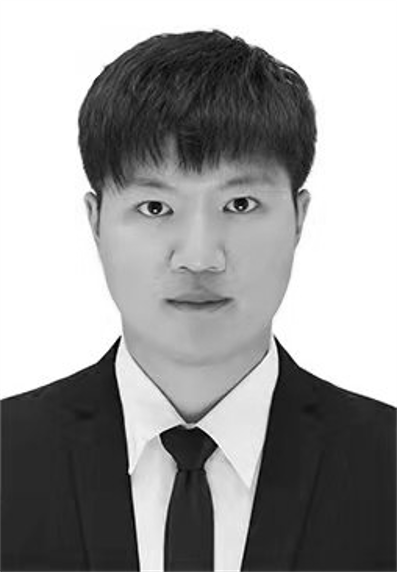}}]{Xin Gao}received his B.S. degree in Computer Science and Technology from China University of Mining \& Technology, Beijing in 2018. He is a Ph.D candidate majoring in Computer Science and Technology in China University of Mining \& Technology, Beijing. His research interests are pattern recognition, multi-modal fusion, image processing.
\end{IEEEbiography}

\begin{IEEEbiography}[{\includegraphics[width=1in,height=1.25in,clip,keepaspectratio]{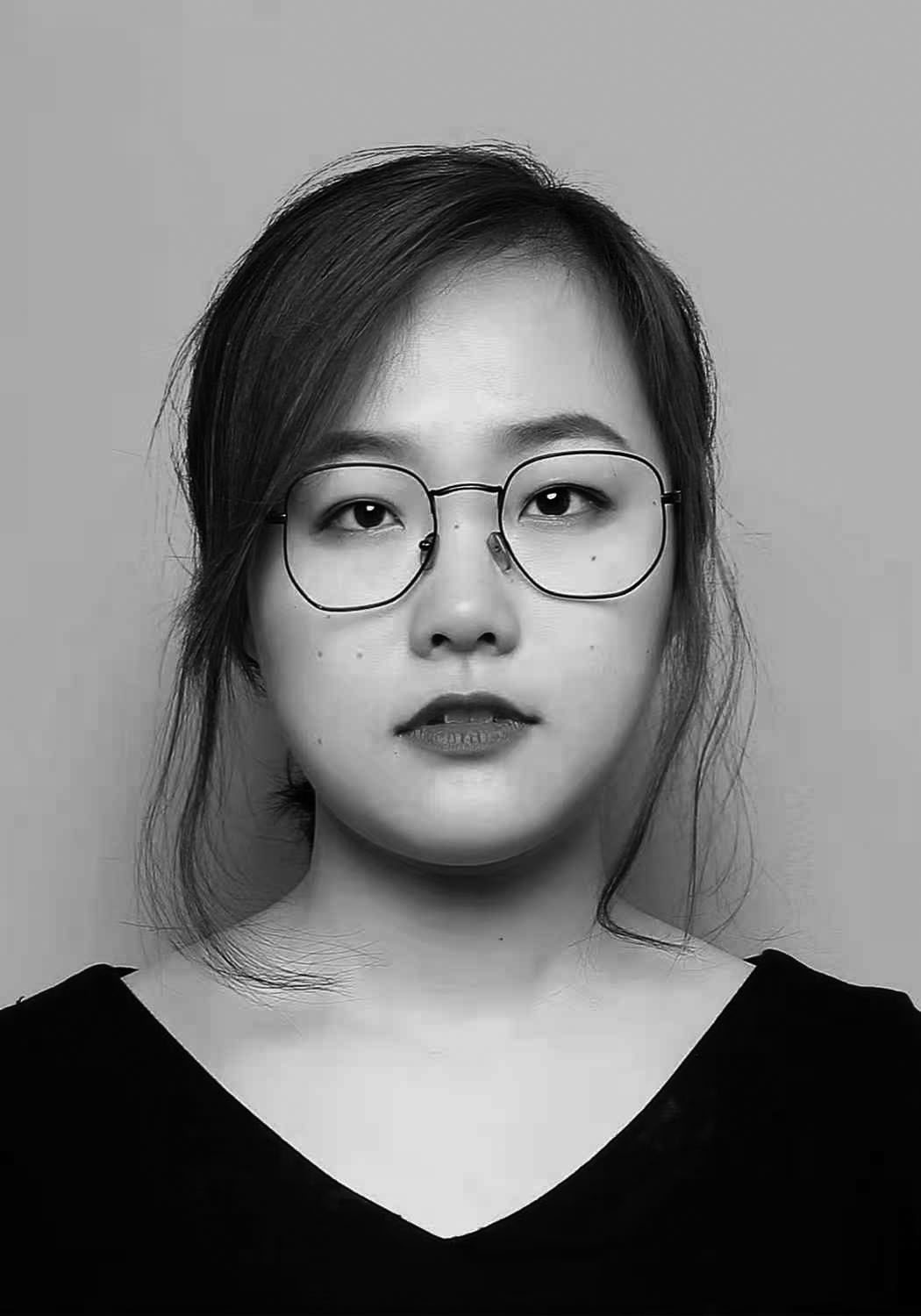}}]{Yijin Xiong}received her B.S. degree from Harbin University of Science and Technology, China, in 2017. In 2018, she joined China University of Mining and Technology (Beijing) to pursue her M.S. degree in computer science, followed by her Ph.D. degree in the Image Research Laboratory, School of Computer Science, China University of Mining and Technology (Beijing). Her current work focuses on multi-sensor information fusion research.
\end{IEEEbiography}

\begin{IEEEbiography}[{\includegraphics[width=1in,height=1.25in,clip,keepaspectratio]{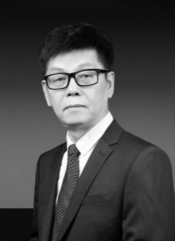}}]{Dafeng Jin} was born in China in 1965. He is an associate professor of the School of Vehicle and Mobility of Tsinghua University. His research field is intelligent driving and integrated technology of new energy vehicle system.
\end{IEEEbiography}

\begin{IEEEbiography}[{\includegraphics[width=1in,height=1.25in,clip,keepaspectratio]{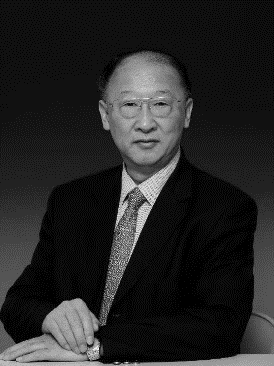}}]{Jun Li} was born in Jilin Province, China in 1958. He received a Ph.D. degree in internal-combustion engineering at Jilin University of Technology, in 1989. He has joined the China FAW Group Corporation in 1989 and currently works as a professor with the School of Vehicle and Mobility at Tsinghua University. Now he also serves as the chairman of the China Society of Automotive Engineers (SAE). In these years, Dr. Li has presided over the product development and technological innovation of large-scale automobile companies in China. Dr. Li has many scientific research achievements in the fields of automotive powertrain, new energy vehicles, and intelligent connected vehicles. Dr. Li is the author of more than 98 papers. In 2013, Dr. Li was awarded an academician of Chinese Academy of Engineering (CAE) for contributions to vehicle engineering.
\end{IEEEbiography}

\begin{IEEEbiography}[{\includegraphics[width=1in,height=1.25in,clip,keepaspectratio]{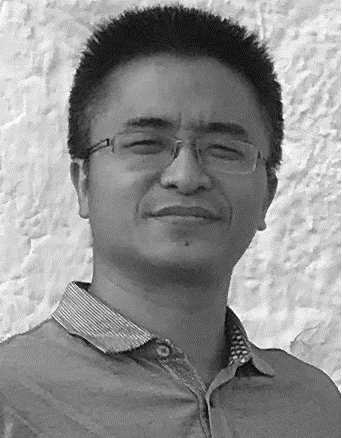}}]{Huaping Liu} received the B.E. degree in mechanical engineering from University of Shanghai for Science and Technology, in 1997, and the M.E. degree in electrical engineering from Tongji University, in 2000. He received the Ph.D. degree in computer science and technology from Tsinghua University, in 2004. Now he is an IEEE senior member. He is currently the associated professor with the Department of Computer Science and Technology at Tsinghua University. He serves as the associated editor for journals, including IEEE Trans. on Automation Science and Engineering, IEEE Trans. on Industrial Informatics, Neurocomputing, and IEEE Robotics and Automation Letters, etc. And he served for the conferences like ICRA, IROS, and IJCAI. Dr. Liu was also a recipient of the Andy Chi Best Paper Award from IEEE Instrumentation and Measurement Society (IMS) in 2017.
\end{IEEEbiography}








\end{document}